\documentclass[lettersize,journal]{IEEEtran}
\IEEEoverridecommandlockouts
\usepackage{colortbl}
\usepackage{cite}
\usepackage{amsmath,amssymb,amsfonts}
\usepackage[font=scriptsize]{caption,subcaption}
\usepackage{algorithmic}
\usepackage{graphicx}
\usepackage{textcomp}
\usepackage{xcolor}
\usepackage{multirow}
\usepackage{pdfpages}
\usepackage{caption}
\usepackage{subcaption}

\begin{document}

\title{\huge{Self-Supervised Learning for Invariant Representations from Multi-Spectral and SAR Images }}

\author{\IEEEauthorblockN{Pallavi Jain, Bianca Schoen-Phelan, Robert Ross }\\
\IEEEauthorblockA{\textit{School of Computer Science} \\
\textit{Technological University Dublin}\\
Dublin, Ireland \\
\{pallavi.jain, bianca.schoenphelan, robert.ross\}@tudublin.ie}
}

\maketitle

\begin{abstract}
Self-Supervised learning (SSL) has become the new state of the art in several domain classification and segmentation tasks. One popular category of SSL are distillation networks such as \textit{Bootstrap Your Own Latent} (BYOL). This work proposes RS-BYOL, which builds on \textit{BYOL} in the \textit{remote sensing (RS)} domain where data are non-trivially different from natural RGB images. Since multi-spectral (MS) and synthetic aperture radar (SAR) sensors provide varied spectral and spatial resolution information, we utilise them as an implicit augmentation to learn invariant feature embeddings. In order to learn RS based invariant features with SSL, we trained RS-BYOL in two ways, i.e. single channel feature learning and three channel feature learning. 
This work explores the usefulness of single channel feature learning from random 10 MS bands of 10m-20m resolution and VV-VH of SAR bands compared to the common notion of using three or more bands. In our linear probing evaluation, these single channel features reached a 0.92 F1 score on the EuroSAT classification task and 59.6 mIoU on the IEEE Data Fusion Contest (DFC) segmentation task for certain single bands. We also compare our results with ImageNet weights and show that the RS based SSL model outperforms the supervised ImageNet based model. We further explore the usefulness of multi-modal data compared to single modality data, and it is shown that utilising MS and SAR data allows better invariant representations to be learnt than utilising only MS data. 

\end{abstract}


\section{Introduction}
\label{sec:intro} 

Computer vision applications to satellite imagery have progressed tremendously over recent years with the availability of multi-modal imagery. Nevertheless, the remote sensing (RS) community faces two major problems: labelled data scarcity and, secondly, dealing with the multi-sensor and multi-modal nature of data obtained from different satellites.

Primarily, RS are broadly divided into Optical and Synthetic Aperture Radar (SAR) sensors. On the one hand, optical sensors provide easy-to-read information and vary in spectral and spatial resolution levels, but are negatively affected by weather conditions, such as clouds. On the other hand, SAR sensors provide high resolution and weather independent data but do not provide readily interpretable images. Over the past decade, computer vision in the RS domain has extended its work from supervised learning to unsupervised learning methods for several tasks in MS and SAR data, such as crop monitoring and assessments \cite{sharma2020machine,ouhami2021computer}, building mapping \cite{zhu2020map,robinson2022fast,kang2022disoptnet}, scene classification \cite{mei2021remote,deng2021cnns}, sea-land segmentation \cite{li2018deepunet}, change detection \cite{khelifi2020deep, chen2021changedetection}, and many more. These works used various deep learning networks such as convolutional neural networks (CNNs), recurrent neural networks (RNNs), attention networks, autoencoders, and GANs to solve RS tasks based on spectral, spatial, and temporal information \cite{cheng2017remote, gong2017feature, pires2019convolutional, ma2019deep, wei2020improved, khelifi2020deep, yuan2021review, garnot2021panoptic, persello2022deep}. Some applications use only MS or RGB data, and some use SAR data, such as flood detection. However, since both have their benefits, several research efforts have investigated the possibility of combining data from these two sensors to obtain better quality information \cite{zhu2017deep,schmitt2017fusion, kulkarni2020pixel, jiang2021deep}. Typically, there are three types of optical-SAR fusion, which are pixel-level, feature-level, and decision-level \cite{zhu2017deep,rudner2019multi3net,kang2022cfnet}. However, simply learning the joint features of both modalities has also shown significant advantages in previous works \cite{ jain2021multi,stojnic2021self}. 
\begin{figure}
    \centering
    \includegraphics[scale=0.33]{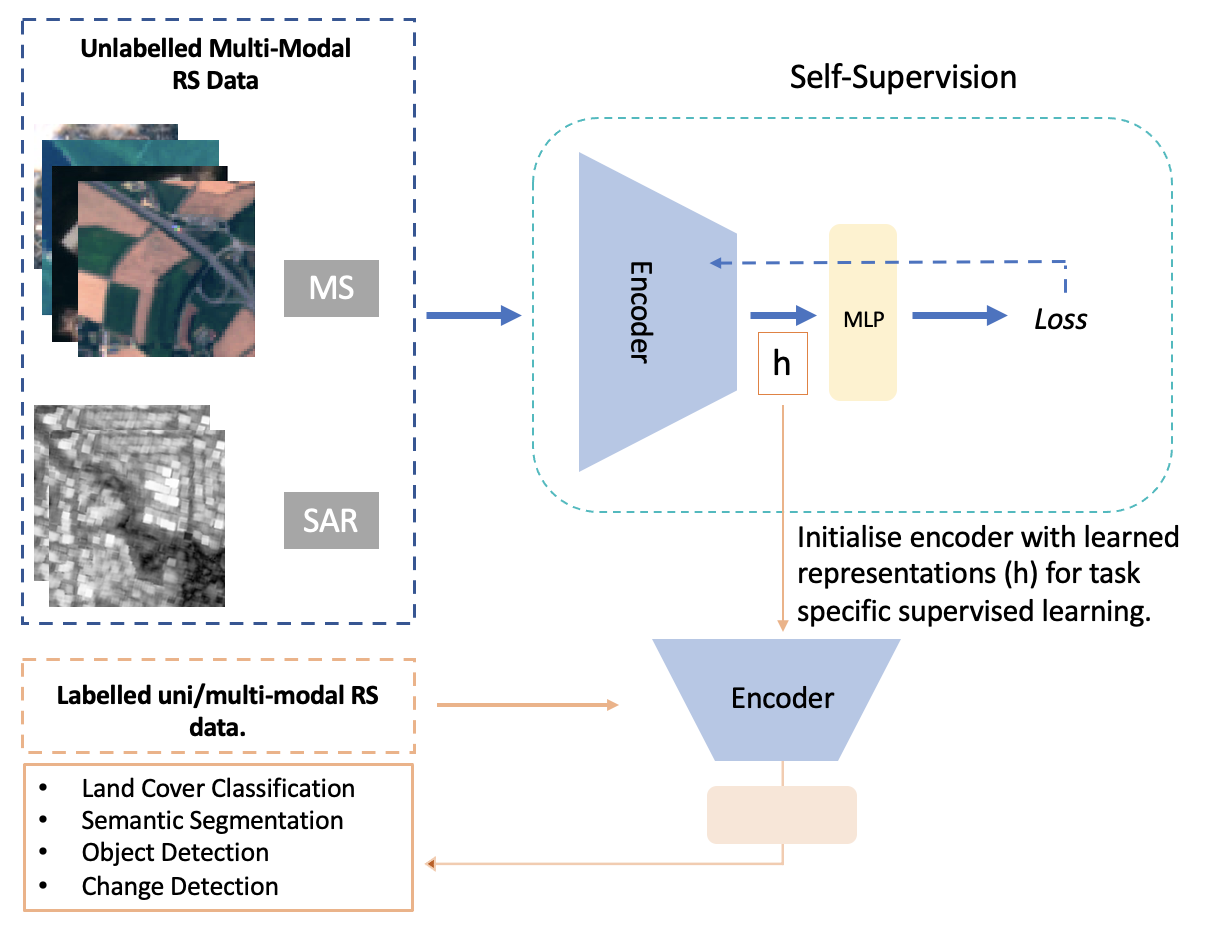}
    \caption{General SSL representation learning approach utilises unlabelled multi-modal data in RS domain. After pre-training, projection layers are removed and encoder weights are transferred for task specific supervised learning.}
    \label{fig:basic_ssl}
    
\end{figure}

Although a considerable amount of work has been performed in the RS domain, and raw satellite data is also freely available, the use of a large amount of labelled data remains a challenge due to the expense of human annotation and domain expertise \cite{li2021image}. To address this problem, transfer learning is one of the solutions that we can leverage from the machine learning community. This, in fact, has shown excellent generalisation capability in several domains \cite{pires2019convolutional,shin2016deep,kim2017end}. Unfortunately, however, most ML pre-trained models are based on RGB and non-RS images, which are different in spectral and spatial aspects from satellite images. This problem has motivated many researchers in the RS domain to obtain a large amount of labelled land cover data to pre-train a supervised model \cite{sumbul2021bigearthnet}, or to apply unsupervised or self-supervised learning to the RS data. With the challenges associated with labelled data, self-supervised learning has become quite popular in the RS community \cite{yuan2020self,vincenzi2021color,kang2021deep,stojnic2021self,jain2021multi}. 

In recent years, self-supervised learning (SSL) has shown competitive results compared to supervised learning \cite{grill2020bootstrap, chen2020simple}. Self-supervised learning utilises the unlabelled data to learn invariant features within the data, and the learnt features are then leveraged for use on downstream tasks with labelled data. In most common SSL architectures, as in Fig. \ref{fig:basic_ssl}, encoders are trained along with the projection layer, and after training, encoder weights are used with a task-specific classifier or decoder. 

SSL can be broadly divided into three categories: pre-text task learning, contrastive learning, and distillation networks. Pre-text task (PTT) learning is based on proxy task prediction such as rotation prediction, solving jigsaw puzzles, colourisation, etc. \cite{gidaris2018unsupervised,noroozi2016unsupervised, zhang2016colorful}. Although PTT learning has shown great advantages in learning different invariant properties, such as rough boundaries and semantic information about the images, it does not generalise well in the latter layers of the network \cite{misra2020self}. Meanwhile, contrastive learning involves the use of positive and negative pairs of images, such as in SimCLR \cite{chen2020simple} or MoCo \cite{he2020momentum}. This usually leads to an increase in the need for memory requirements and an increase in the computational cost since they depend on a high batch size for the contrastive approach to be most effective. Recently, distillation networks such as BYOL \cite{grill2020bootstrap} and DINO \cite{caron2021emerging} have shown the potential of using only positive pairs to learn better representations, ultimately reducing the need for a large batch size compared to other contrastive learning approaches. Although these methods have claimed to add several factors to models that should prevent the common collapse problem in Siamese networks, this is still an active area of research. 

In SSL type learning, apart from network architectures, the design of data augmentation such as random crops, random rotation, colour jitters, random greyscale, etc., is a crucial part of the learning process. These augmentations preserve the semantic aspect of the data and provide better invariant representations \cite{von2021self}. The goal of distillation networks such as BYOL is to learn invariant feature embedding by bringing these augmented views closer. Recent work in the RS domain with SSL is being used for classification, segmentation, or change detection tasks, yet there are very few works which utilise the multi-modality of the remote sensing domain \cite{chen2021selffusion, wang2022self}. We argue that having the use of multi-modal satellite data can provide implicit augmentation to the model when we use MS and SAR data as augmented pairs of images. Motivated by this, we trained a remote sensing-based SSL distillation network (RS-BYOL) against the MS and SAR data to learn invariant spectral and spatial features. Although both data sources vary in terms of resolution and modality, our hypothesis is that feature-level information should remain the same and provide valid inputs to the distillation network process. 

In summary, the main contributions of this work are then as follows:
\begin{itemize}
\item We investigate the importance of having RS-based pre-trained models compared to traditional RGB or non-RS based pre-trained models. For this, we compared our BYOL based RS-BYOLs weights with popular ImageNet and MS COCO pre-trained weights. We also compare our results with another RS-based contrastive learning model, SeCo \cite{manas2021seasonal}. This further investigates the suitability of self-supervised distillation networks for RS-based learning. 
\item We explore the usefulness of multi-modality for feature learning rather than learning by a single modality. For that, we train RS-BYOL with MS-SAR data and, also, with only MS data, which further confirms the effectiveness of learning invariant features from MS-SAR data.
\item Finally, we also validate single-band feature learning efficiency against three-band feature learning in satellite imaging. In general, each MS band has its own reflectance property to differentiate between geographical features, such as water, land cover, and more. This motivates us to explore single-band invariant feature learning, which we believe can learn powerful features through a model combining three or more bands.
\end{itemize}


\begin{figure*}
  
    \centering
    \begin{tabular}{cc}
        \subfloat[]{\includegraphics[width=0.54\textwidth]{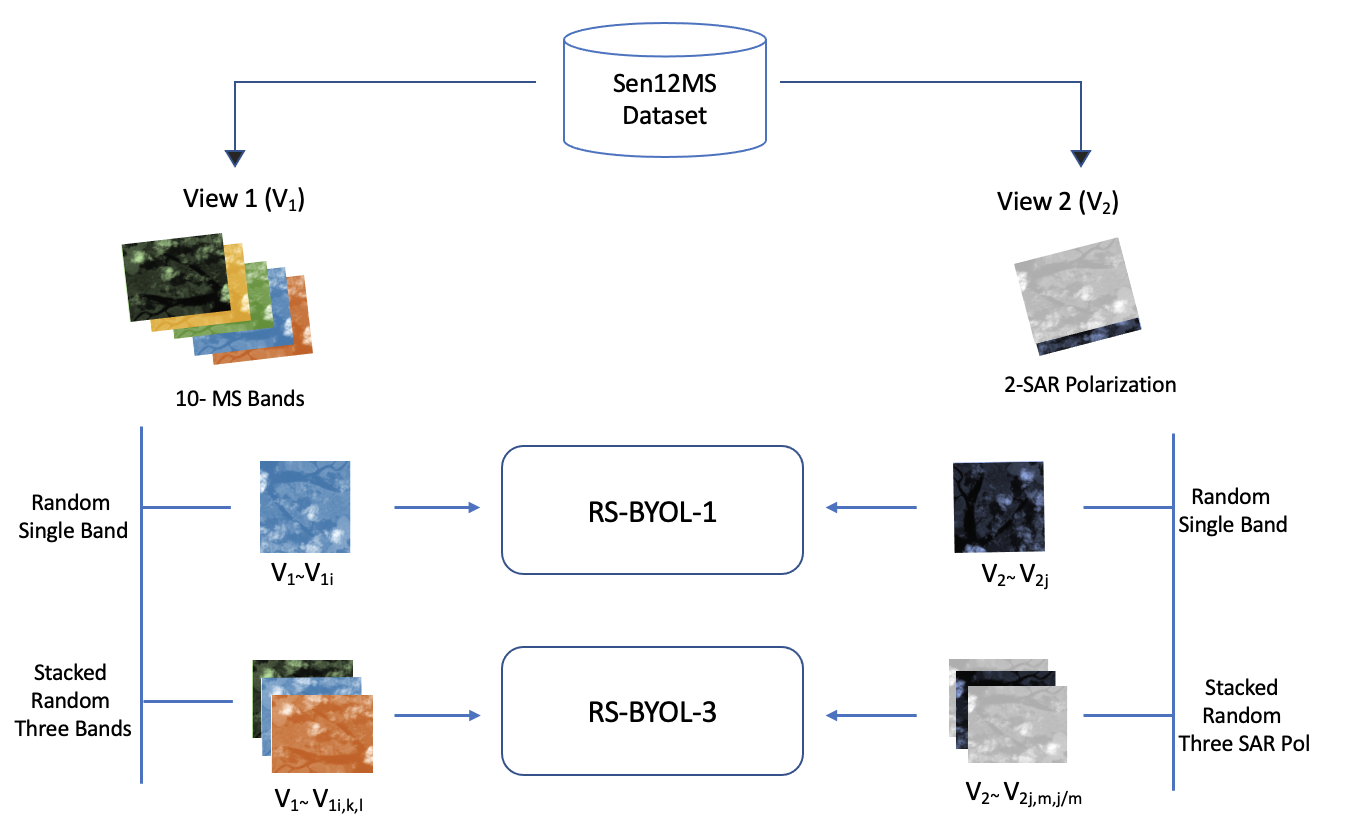}} 
        \subfloat[]{\includegraphics[width=0.43\textwidth]{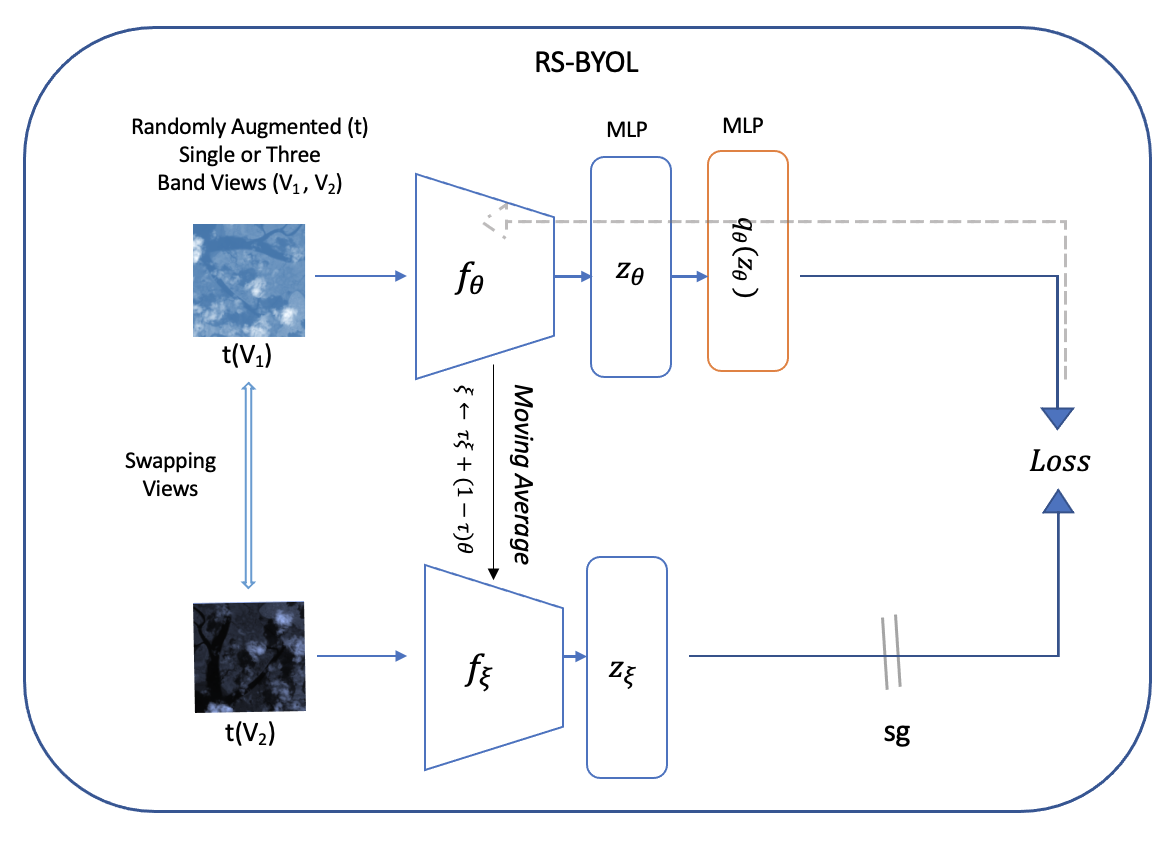}} 
    \end{tabular}
    \caption{(a) MS bands and SAR polarisation are two views $V_1$ and $V_2$ respectively, where single channel model utilises random single bands and three channel model utilises three bands stacked randomly. We utilised 10 MS bands of 10m-20m resolution and discarded 60m atmospheric bands, for SAR we utilised both polarisation, where $i$$\neq$$j$$\neq$$k$ and represents random MS bands while stacking for three channel whereas $j$,$m$ represents two polarisation in random order with third as repetitive of either $j$,or $m$. (b) BYOL architecture with RS data, which consist of an Encoder ($f$), an MLP projection head ($z$), and an MLP prediction head ($q(z)$), \textit{sg} represents stop gradient, ${\theta}$, and ${\xi}$ are different weights, where ${\xi}$ is updated based on ${\theta}$ and ${\tau}$ . Input images do not represent actual band here, but used only for multiple band representation.}
    \label{fig:model_arch}
\end{figure*}

\section{Related Work}
\label{sec:lr}
This section discusses previous relevant work on self-supervised learning and its application in the remote sensing domain.

\subsection{Optical-SAR Fusion and Computer Vision}
Computer vision in the remote sensing domain has progressed greatly in both classification, object detection, and segmentation tasks. 
With that, the number of applications in remote sensing has also increased and is not only limited to scene classification, but also extends to crop monitoring \cite{zhao2020augmenting, qiao2021crop}, flood detection \cite{akiva2021h2o}, or building detection \cite{rudner2019multi3net, bischke2019multi} and many more. Considering these applications, many supervised and unsupervised learning methods have been proposed over the years, and extensive use of CNNs, attention networks, autoencoders, and GANs has been seen \cite{cheng2017remote, gong2017feature, pires2019convolutional, ma2019deep, wei2020improved, khelifi2020deep, yuan2021review, garnot2021panoptic, persello2022deep}. 

Each band in MS has its own characteristics, e.g., red-edge bands that are good at detecting vegetation and soil \cite{clevers2013remote, liu2021comprehensive}, SWIR and NIR bands that are good at detecting water and parsing thin clouds \cite{ceccato2001detecting,wang2006cloud,wang2013remote}, while SAR provides images regardless of weather conditions, which helps in the study of terrain or disaster related tasks \cite{landuyt2018flood,zhang2018polarimetric}. These properties of MS and SAR bands are utilised in many applications such as crop or weed detection, flood detection, or oceanography, which requires more than just an RGB image. Although SAR has its benefits and provides additional information missed in MS, it is highly susceptible to speckle noise, which encouraged researchers to use them jointly. The optical-SAR fusion technique is divided into three levels: these are the pixel level, feature level, and decision level. Where pixel-level fusion is computationally costly and requires a strict registration technique, feature-level fusion uses feature extraction, where feature maps of optical and SAR data are combined, decision-level fusion is less sensitive to registration and combines individually classified images \cite{klein1999sensor,kulkarni2020pixel}. Many works showcased the different efficiencies of these fusion techniques, such as Rajah et al. who showed a feature-level fusion framework for the detection and mapping of plant species, while Zhang et al. used a framework for classification of land cover in mountainous areas \cite{zhang2020novel}. Multi3Net by Rudner et al. demonstrated pixel-level and decision-level fusion for flooded building mapping \cite{rudner2019multi3net}. All these works showed that fusing optical and SAR images is advantageous in comparison to using a single modality. However, we believe that looking at optical and SAR images as augmented counterparts and learning joint representations is potentially an excellent source for feature learning. 

\subsection{Contrastive Learning and Distillation Networks}
Self-supervised learning (SSL) performance has closed the gap with supervised learning \cite{grill2020bootstrap,caron2021emerging}, which has helped solve the big problem of labelled data requirements for deep learning. 

Many contrastive and distillation networks have been proposed with increasing interest and performance enhancements. These networks are based on the general class of Siamese networks, which learn feature embedding based on the similarity between two images. Unfortunately, the early Siamese networks often suffered from model collapse to trivial solutions \cite{grill2020bootstrap,chen2021exploring}. However, with careful control over the input data, model architecture, and loss function, contrastive and distillation network approaches have been able to avoid this collapse. 

Contrastive learning, such as SimCLR \cite{chen2020simple}, deals with positive and negative pairs of images, with the intention of maximising the agreement in projections between similar images while minimising for dissimilar images. Similarly, MoCo \cite{he2020momentum} leverages the concept of a momentum encoder and a memory bank for negative feature vectors. Although these learning methods show good performance, they are computationally costly due to the use of negative pairs that require large batch sizes and memory requirements. 

Meanwhile, distillation networks such as BYOL \cite{grill2020bootstrap} and DINO \cite{caron2021emerging} use only positive pairs and have become quite popular in terms of performance and memory efficiency. BYOL and DINO are similarity-based networks whose architecture is inspired by concepts from MoCo \cite{he2020momentum} and follow a student-teacher approach. Here, teacher encoders do not have error-signal backpropagation; instead, they utilise slow moving averages from the student network weights. This makes them asymmetric and prevents the embeddings from collapsing. 

In addition to contrastive and distillation networks, Barlow Twin (BT) \cite{zbontar2021barlow} has been proposed, which employs a redundancy reduction approach. This model architecture is similar to SimCLR except that it utilises only positive pairs and the loss is calculated based on a cross-correlation matrix between the projections. BT shows competitive results with BYOL on certain aspects, such as batch size and asymmetry requirements, but BT networks are not robust toward augmentation and require higher projection output. Although this network trains well with a smaller batch size (1024) compared to BYOL, this still requires us to train both sides of the network. This ultimately makes it more computationally costly, where hardware availability is still a constraint. 

Although continuous improvement has been seen in the architectures of SSL networks, this area is still being explored thoroughly, with efforts focusing on examining each network layer, the usefulness of stop gradient in collapsing, and investigation of appropriate loss functions. As all network performance fluctuates across different datasets and with different parameters and network settings, it is not yet possible for us to determine which of these models ultimately performs best. 

\subsection{SSL and Remote Sensing}

Due to the lack of available labelled data in the remote sensing (RS) domain, and the gap which exists with non-RS RGB images in terms of spectral and spatial context, work in the RS domain is now leveraging self-supervised learning to pre-train RS based models for downstream tasks.

Recent literature has shown the popularity of applying both pre-text tasks (PTT) and contrastive learning methods. The work by Vincenzi et al. \cite{vincenzi2021color} utilises colourisation as a pre-text task to learn spectral information using multi-spectral images. Meanwhile, the work by Tao et al. applies instance discrimination-based PTT to RGB remote sensing data, which performs well in downstream tasks with small datasets \cite{tao2020remote}.

Some works use contrastive learning with the application of MoCo based models to temporal \cite{ayush2021geography} and spatially augmented RGB remote sensing images \cite{kang2021deep}. SeCo\cite{manas2021seasonal} is one such approach, which utilises the MoCo architecture with the multi-augmentation contrastive learning method taking into account the seasonal contrast in the RGB remote sensing data. Also, quite a few works have attempted to fully exploit both sensor modalities, i.e., multi-spectral and SAR images, for representation learning with contrastive or distillation networks. One such work uses contrastive multiview coding (CMC) by leveraging multiple high- and low-resolution remote sensing datasets \cite{stojnic2021self}. Chen \& Bruzzone shows the use of both modalities using contrastive loss for pixel-level learning \cite{chen2021selffusion}. Another work, by Chen \& Bruzzone, concatenates MS and SAR temporal images to obtain pixel-level discrimination embedding with the distillation network for change detection \cite{chen2021changedetection}. Meanwhile, work by Jain et al. demonstrates that single channel features can also provide significant advantages in learning invariant representations for satellite data \cite{jain2021multi}. Similar results are shown by Montanaro et al. \cite{montanaro2021self}, which uses a combination of contrastive learning and the text task with single-channel feature learning. Though it has been seen that similarity-based networks such as SimSiam do not perform well in the case of remote sensing \cite{jain2021multi,montanaro2021self}, we believe that distillation networks provide additional advantages for the learning of invariant spatial and spectral representations. 

In addition, all these works highlighted the benefit of learning RS based representations for many downstream tasks rather than utilising RGB natural image based pre-trained models. Moreover, those works have shown that SSL has been established as providing a solution to the problem of large labelled data without compromising performance compared to supervised learning. This motivates us to provide better experimental analysis of self-supervised distillation networks for RS representation learning by utilising both MS and SAR data. 


\section{Methodology}
This section presents our framework for multi-modal satellite image representation learning based on distillation networks in self-supervised learning. This framework builds on the concept of distillation networks and, in particular, the BYOL architecture, but adopts a particular strategy to the application and training process for multi-modal data. 

Self-supervised learning is based on the weak smoothness assumption, which states that distortion or the addition of perturbation to an image does not change its labels \cite{van2020survey}.  This work revolves around the same concept with multi-view representation similarity. To put this in another way, since MS and SAR images are two views of the same location, our modelling leverages the implication that having different sensor images does not change the images' geographical properties. 

Let us assume that the MS and SAR images are in practice two views $V_{1}$ and $V_{2}$, and that $f(V)$ is a representation of a view. The goal of this work is then to increase the similarity between the representations of the two views, say: 
\[f(V_{1}) \Leftrightarrow f(V_{2})\] \label{eq:eq_rep} 
As indicated, the most appropriate network for such learning is arguably the siamese networks due to the duality of the data. However, the trivial solution to siamese networks is prone to collapse to constant representations, which led to the rise of contrastive and distillation networks. 

Considering that our goal is to learn invariant representations of MS ($V_1$) and SAR ($V_2$), we trained a remote sensing network (RS-BYOL) based on the BYOL architecture. The architecture follows the Student-Teacher type asymmetry, where students and teachers have identical encoders ($f(T(V)$). Furthermore, the weights are updated only for the student side of the network, whereas the weights of the teacher network are modified based on the moving average of the student weights. In order to add more asymmetry to the network, an extra predictor, i.e. a multi-layer perceptron (MLP) head, has been added to the student side of the network. The network compares the L2 normalised projection from the student and teacher sides of the network, which is an output of a predictor head ($q(z)$) and a projection head ($z$) respectively. The projection and prediction heads are 2-layer multi-layer perceptrons (MLP) of hidden projection size 256 with batch normalisation and a non-linear (ReLU) layer. 

The loss is calculated between the L2 normalised prediction head output $q(z_{\theta})$ from the student side and the projection head output $z_{\xi}$ from the teacher side. Here ${\theta}$ is the weight of the student network, ${\xi}$ is the weight of the teacher network, which is an exponential moving average of $\theta$. As shown in Eq. \ref{eq:xiupdate}, $\xi$ is updated by $\theta$ and the decay rate ($\tau$), which in this network is 0.9.
\begin{equation}
        \xi \leftarrow \tau \xi + (1-\tau)\theta
        \label{eq:xiupdate}
 \end{equation}
This implicitly provides asymmetry and prevents the model from learning the same latent representations, which in turn prevents collapse. 

The loss function for the network is represented as in Eq. \ref{eq:loss}, where $\left \|\cdot \right \|_{2}$ is the normalisation of $L_2$, and $t$ and $t'$ are the augmentations applied to the views $V_1$ and $V_2$:

\begin{equation}
   L(t(V_1),t'(V_2)) = 2-2\cdot \frac{q(z_{\theta,1})}{\left \| q(z_{\theta,1}) \right \|_{2}}\cdot\frac{z_{\xi,2}}{\left \| z_{\xi,2} \right \|_{2} }
    \label{eq:loss}
\end{equation}


Our loss function is the mean square error (MSE) between the two projections for the augmented views of the MS and SAR images. The total loss is calculated by interchanging views between the student and teacher network, which results in the total MSE as in Eq. \ref{eq:totalloss}.
\begin{equation}
        Loss_{total} = L(t(V_1),t'(V_2)) + L(t'(V_2),t(V_1)) \label{eq:totalloss}
 \end{equation}

This architecture not only prevented the collapsing issue, but also provided better performance than some of the contrastive learning approaches and helped reduce the batch size requirement for training. Although this architecture performs well, many research works have shown that different analyses, such as the use of the batch normalisation layer, do not help to avoid collapse \cite{richemond2020byol}, but rather that the stop gradient is the key to avoiding collapse \cite{chen2021exploring}. In other words, these works show that there is still a need for more analysis in terms of the reasons for the outstanding performance of BYOL-like distillation networks and further need to explore them in other domains such as remote sensing.

In order to better understand the representation learnt from the RS data, we pre-trained our models into two variants: (i) a single-channel model, and (ii) a three-channel model. For the single channel model, two views were randomly selected from the MS and SAR bands, that is, one MS band ($V_{1i}$) out of 10 bands, i.e. $i = \{1,2...10\}$, and one SAR polarisation ($V_{2j}$), i.e. $j = \{1,2\}$. The three-channel model utilises the same approach, except that it draws upon a random three bands from the MS data and generates three-channel SAR images by randomly copying a third channel from the two polarisations. These two views are then randomly augmented with the transformations $t$ and $t'$ before being sent to the encoder. 

After pre-training of the model, MLP heads are removed, and we use only student encoder weights for the downstream task, which in our case we evaluate on different remote sensing datasets.

This work demonstrates the usefulness of RS-BYOL, which is trained to learn single and three channel RS-based invariant features. We compare our work with models trained from scratch, random initialisation, ImageNet \& MS COCO  pre-trained ResNet50, BYOL-ImageNet \cite{yaox12}, and SeCo \cite{manas2021seasonal} weights to explore the usefulness of RS-BYOL models against models trained with non-RS data and RS-based contrastive models. For BYOL-ImageNet pre-trained weights, we utilised the weights published by Yao Xin \footnote{https://github.com/yaox12/BYOL-PyTorch} \cite{yaox12}, the author trained the BYOL architecture with a ResNet50 encoder, with batch size of 128 and 300 epochs. 

\section{Experiment}
In this section, we discuss the implementation details of the RS-BYOL models and give details on the datasets and data augmentation used.

\subsection{Datasets}
We pre-trained our models with the Sen12MS dataset \cite{schmitt2019sen12ms}, which consists of 180,662 triplets of Sentinel-1, Sentinel-2 image patches and MODIS land cover map images of size 256 x 256 pixels. Whereas patches from Sentinel-1 are a dual-polarisation synthetic aperture radar (SAR), and have VV and VH polarisation, Sentinel-2 has 13 multi-spectral (MS) bands. Among the 13 MS bands, in this work, we used 10 bands of 10m-20m resolution and discarded three atmospheric bands of 60m resolution, and for VV and VH, we used both polarisations. We utilised 50\% of the data for our work -- that is 90K pairs of images. The reason for using 50\% data is due to hardware limitation, as a larger dataset adds computational cost.

For evaluation of the representations learnt by the models, we utilised two benchmark datasets for a land cover classification task, i.e., EuroSAT\cite{helber2019eurosat} and RESISC45\cite{cheng2017remote}. EuroSAT consists of 64 x 64 size images and is similar to Sen12MS in terms of sensor and spectral information. It consists of 10 land cover classes with 27,000 images. Meanwhile, the RESISC45 dataset consists of high-resolution RGB images of 256 x 256 pixels in size and contains 31,500 images, covering 45 scene classes. We also evaluated our models on 5000 images from Sen12MS.

Finally, we use the 2020 IEEE Data Fusion Contest (DFC) land cover mapping dataset \cite{sen12ms} for segmentation evaluation, which is a subset of Sen12MS data and contains 12,228 pairs of labelled Sentinel-1/2 data. Of these, we used 900 images for the linear probing evaluation. The dataset consists of 10 land cover classes based on the IGBP classification scheme. Among the 10 classes, we excluded ``Savanna" and ``Snow/Ice", as these classes are rare in this subset of Sen12MS, and utilised 8 land cover classes, that is, forest, scrubland, grassland, wetlands, croplands, urban / built-up, woodland, and water.

\begin{figure}
\centering
\begin{tabular}{c}
\subfloat[EuroSAT Full]{\includegraphics[width=0.45\textwidth]{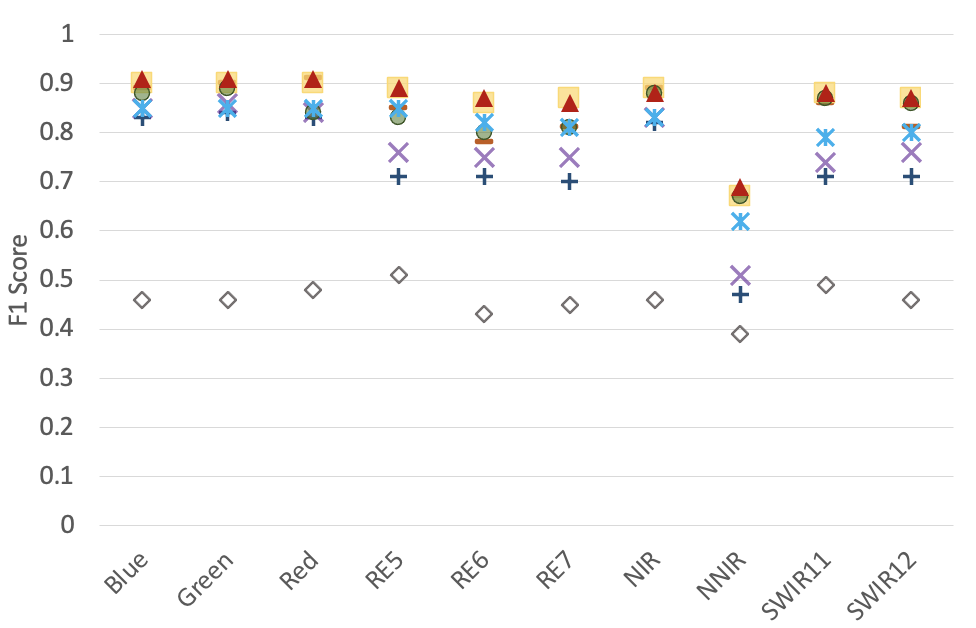}} \\
\subfloat[EuroSAT 7K]{\includegraphics[width=0.45\textwidth]{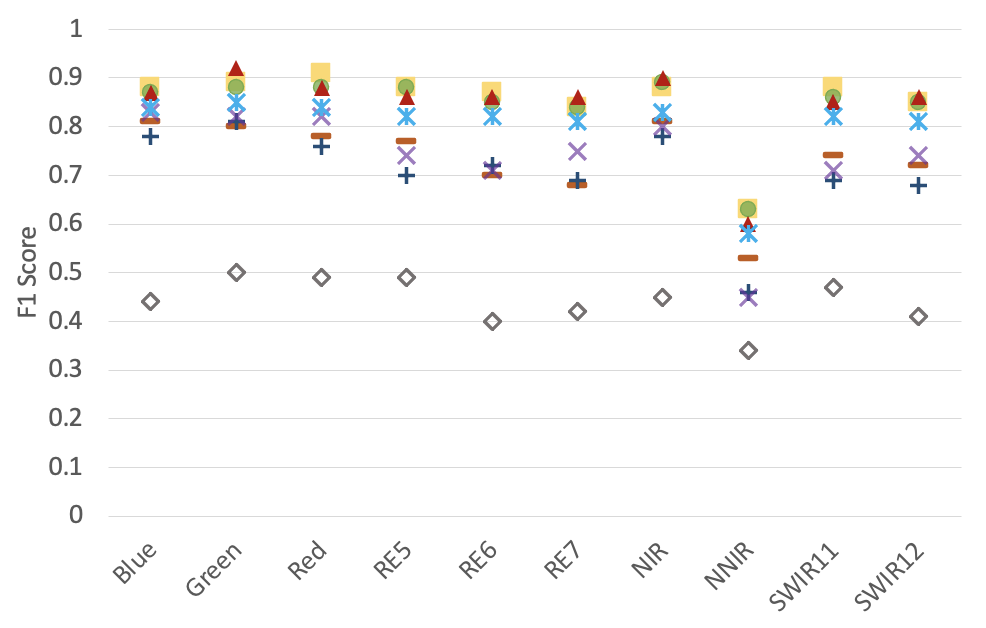}} \\
\subfloat[Sen12MS]{\includegraphics[width=0.45\textwidth]{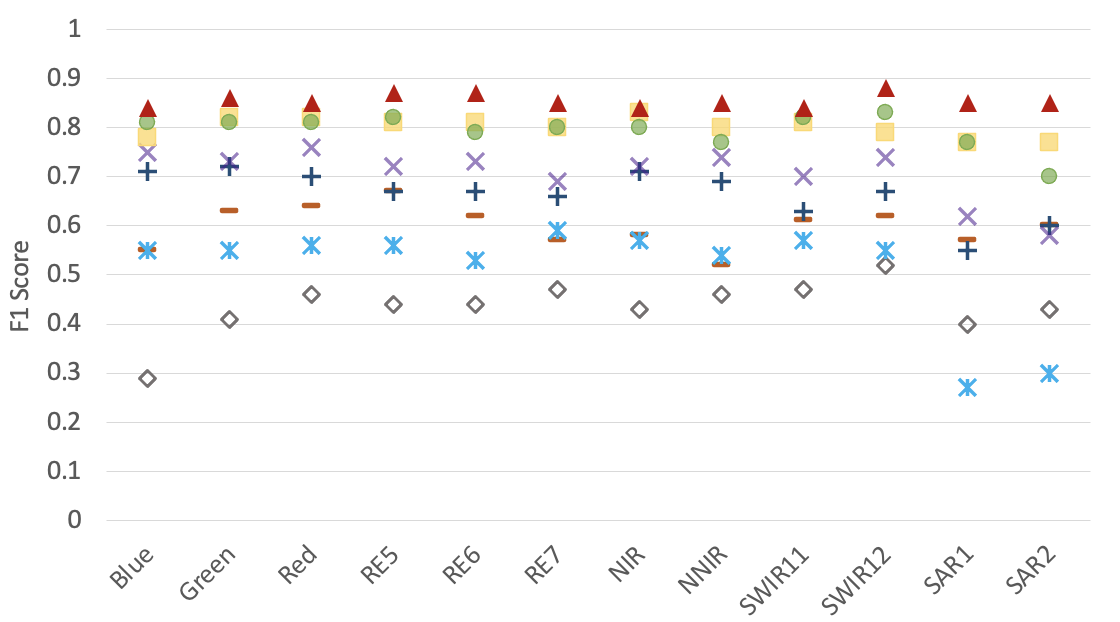}} \\
\subfloat{\includegraphics[width=0.45\textwidth]{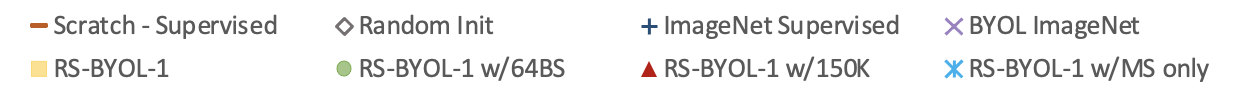}} 

\end{tabular}
\caption{Band-Wise Linear Probing (F1 Score) for Single Channel Models on EuroSAT and Sen12MS Datasets}
\label{tab:evalchannelwise}
\end{figure}

\subsection{Data Augmentation}

\begin{figure}  
    \centering
    \begin{tabular}{c}
        \subfloat[MS]{\includegraphics[width=0.4\textwidth]{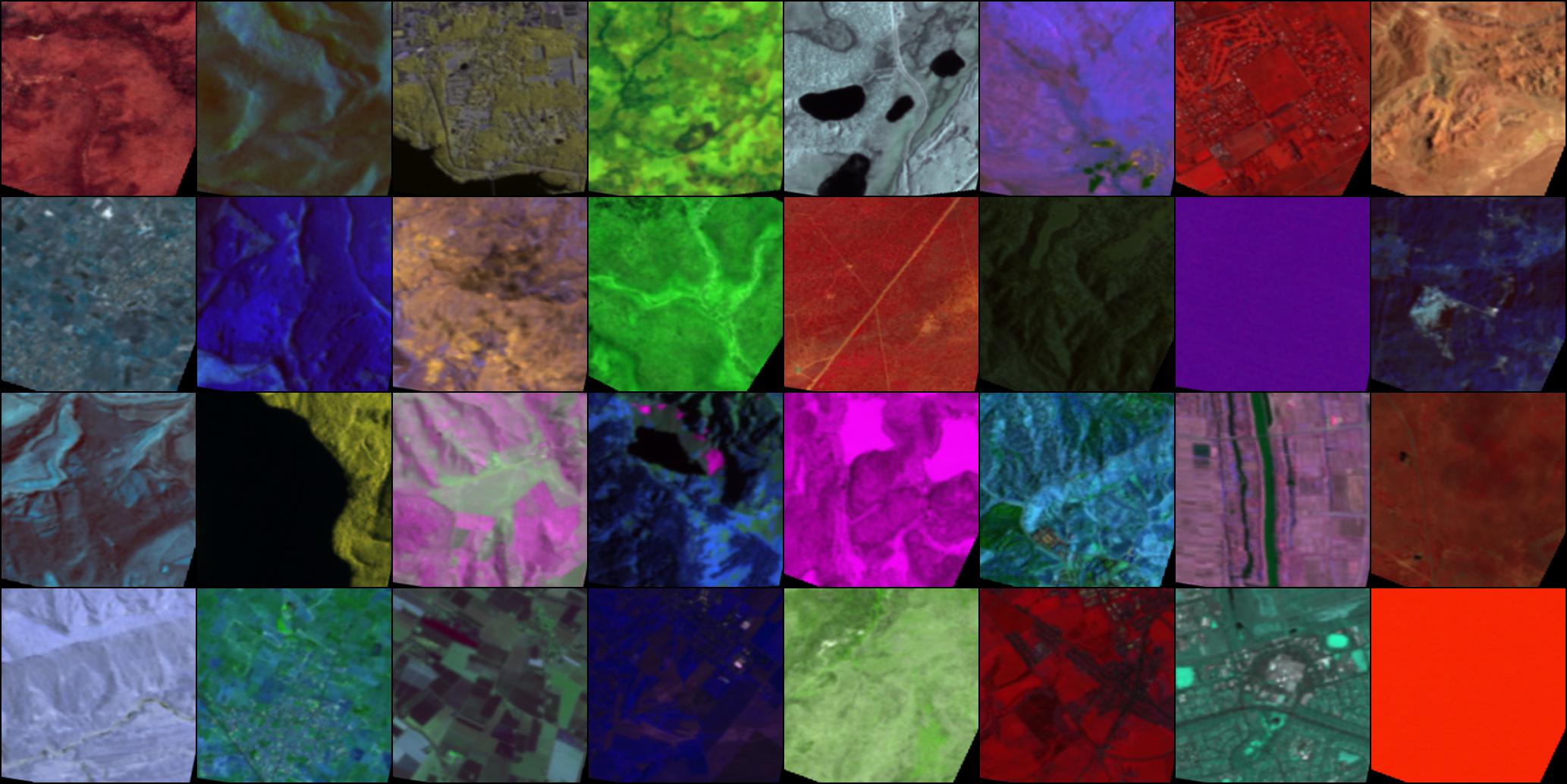}} \\
        \subfloat[SAR]{\includegraphics[width=0.4\textwidth]{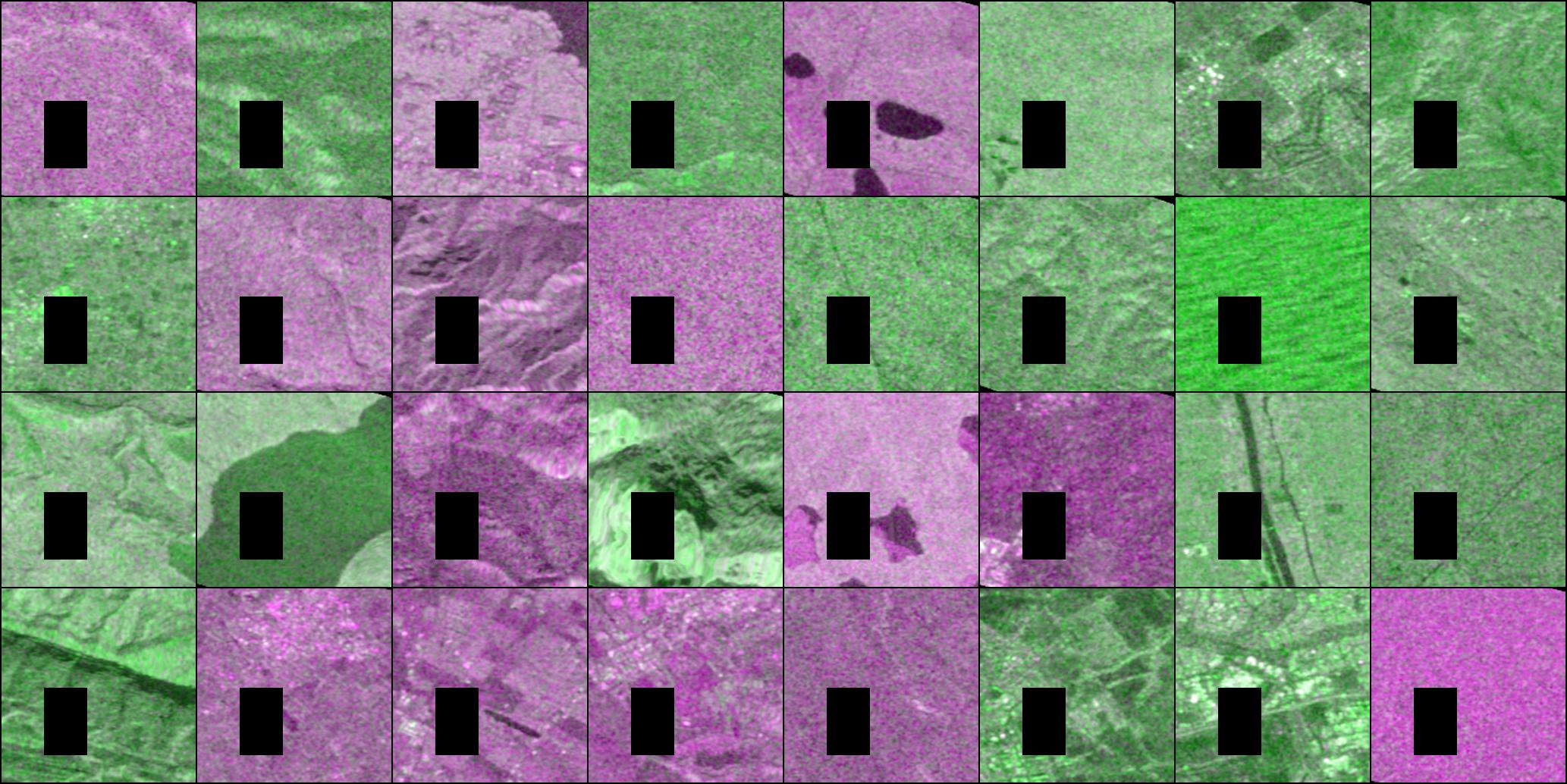}} 
    \end{tabular}
    \caption{Example: A batch of randomly augmented three channel MS and SAR images.}
    \label{fig:aug}
\end{figure}

We used all the augmentation techniques applied in BYOL \cite{grill2020bootstrap}, except colour jitter, as this work implicitly applies colour jitters to images through the use of random MS and SAR bands. We also utilised random pixel erasing to provide a stronger augmentation to the images. Thus, our augmentations include random flips, rotation, Gaussian blur, random pixel erasing, and resized crops with bicubic interpolation. We also applied random greyscale to three-band images. An example of a batch of randomly generated augmented images is shown in Fig. \ref{fig:aug}.

\subsection{Experimental Settings and Parameters}
For pre-training the model, we used ResNet50 as the backbone encoder and the SGD optimiser with a cosine warm restart scheduler. The warm restart of 10 epochs was given with a multiplier of 2, and the learning rate ranges between 0.002 and 0.2. We utilised a batch size of 32 and trained all our models for 400 epochs with weight decay of 1e-5. 

The following subsections explain individual settings and variations for single channel and three channel model variants.

\subsubsection{Single Channel Models}
For single channel feature learning, RS-BYOL-1 was trained with random bands from MS as view $V_1$ and a random selection of VV and VH from SAR as view $V_2$. 

We trained RS-BYOL-1 in multiple variations to look at the impact of each RS-BYOL's representation learning. These variations were: (i) increasing batch size, (ii) increasing dataset size, and (iii) utilising only the multi-spectral dataset.

Most of the best performing SSL networks such as BYOL, DINO, and MoCo utilise much larger batch sizes than 32 to attain the maximum performance; thus, we also trained RS-BYOL-1 with 64 batch size. This variation was made to analyse the impact of batch size on feature learning. As our dataset is not as large as the datasets on which typically large networks are trained on, making use of a large batch size would likely not be a benefit for RS-BYOL-1 in practise. This upper batch size of 64 was influenced by our hardware limitations.

Furthermore, we analyse the impact of training with different dataset sizes on RS-BYOL-1. For this training, we increased the size of the dataset by 30\% using 150K pairs of Sen12MS. 

Finally, to better understand the usefulness of SAR data in learning invariant representation, we also trained our model with a single modality, that is, MS data only. This study examines the benefit of using multi-sensor data for remote sensing representation learning.

Each of these variations was trained with the same architectural and parameter settings, that is, encoder, optimiser, scheduler, epochs, and batch size, except for the case of the 64 batch size variation.

\subsubsection{Three Channel Models}

For the three-channel remote sensing based feature learning model, we utilised two different input techniques. The first technique, RS-BYOL-3, is similar to our single-channel model (RS-BYOL-1), which takes three random MS bands as one view ($V_1$) and SAR's VV \& VH bands as another view ($V_2$); we randomly copied the third channel for the third SAR view. Another input technique which we applied is \textit{RGB-S1S2}, which keeps one view ($V_1$) as red, green, and blue bands (RGB), and another view ($V_2$) with one SAR and two random non-RGB MS bands such as red edge, narrow infrared or shortwave infrared bands. 

We utilised a single Nvidia RTX 2080Ti GPU for pre-training and evaluation of the models.


\begin{table*}[!htb]
\centering
\caption{Channel-wise Linear Probing on DFC Segmentation Dataset}
\resizebox{18cm}{!} {
\begin{tabular}{c|c|ccccccccccccc} 
\hline
\multicolumn{2}{c|}{Band}                       & \multicolumn{1}{l}{Blue} & \multicolumn{1}{l}{Green} & \multicolumn{1}{l}{Red} & \multicolumn{1}{l}{RE5} & \multicolumn{1}{l}{RE6} & \multicolumn{1}{l}{RE7} & \multicolumn{1}{l}{NIR} & \multicolumn{1}{l}{NNIR} & \multicolumn{1}{l}{SWIR1} & \multicolumn{1}{l}{SWIR2} & \multicolumn{1}{l}{SAR1} & \multicolumn{1}{l}{SAR2} & \multicolumn{1}{l}{Average}  \\ 
\hline
\multirow{2}{*}{RS-BYOL-1~}       & mIoU           & 54.8                     & 57.7                      & 57.8                    & 58.3                    & 54.5                    & \textbf{56.6}           & \textbf{57.6}           & 55.5                     & 58.1                       & \textbf{58.5}              & 57.2                     & 57.2                     & 56.98                        \\
                               & AA & 68.0                     & 72.2                      & 70.7                    & 72.1                    & 68.2                    & \textbf{69.7}           & 70.8                    & 68.4                     & 71.4                       & \textbf{72.3}              & 70.1                     & 70.5                     & 70.38                        \\ 
\hline
\multirow{2}{*}{RS-BYOL-1 w/150K} & mIoU           & 54.3                     & 58.6                      & 57.6                    & 57.6                    & 56.4                    & \textbf{56.6}           & \textbf{57.6}           & 55.3                     & 57.6                       & 57.9                       & 57.5                     & 57.5                     & 57.05                        \\
                               & AA & 71.0                     & 72.1                      & \textbf{73.0}           & 70.7                    & 69.8                    & 69.6                    & \textbf{71.2}           & 67.9                     & 70.6                       & 71.5                       & \textbf{71.2}            & \textbf{71.8}            & 70.87                        \\ 
\hline
\multirow{2}{*}{RS-BYOL-1 w/64BS} & mIoU           & \textbf{57.7}            & \textbf{59.6}             & \textbf{58.8}           & \textbf{59.1}           & \textbf{57.5}           & 54.8                    & 56.0                    & \textbf{56.5}            & \textbf{59.2}              & 58.4                       & \textbf{58.3}            & \textbf{57.8}            & \textbf{57.81}               \\
                               & AA & \textbf{72.0}            & \textbf{72.7}             & 72.5                    & \textbf{73.1}           & \textbf{71.3}           & 68.0                    & 69.6                    & \textbf{69.6}            & \textbf{72.4}              & 72.0                       & 70.9                     & 71.2                     & \textbf{71.27}               \\
\hline
\end{tabular}
}
\label{tab:channelwisesegresults}
\end{table*}
\section{Evaluation}
For the evaluation of RS-BYOL's learnt representations, we performed a linear probing. That is, we froze the encoder weights and added a simple classifier head. The classifier head was an MLP with a hidden dimension of 128, followed by batch normalisation layers, ReLU, and a softmax layer. We trained all the model variants for 100 epochs and utilised cross-entropy loss as our loss function. For downstream classification tasks we compared our RS-BYOL weights with models trained from scratch, random initialisation, and ImageNet pre-trained ResNet50 and BYOL weights for both single channel and three channel models. We also compared three channel RS-BYOL weights with SeCo weights, which are trained on 1 million RGB remote sensing images with a contrastive learning architecture.

Additionally, in order to evaluate how well representations learnt pixel level information, we also provided an evaluation on land cover mapping with the DFC 2020 Sen12MS dataset. Both single and three channel RS-BYOL utilised the DeepLabV3 model architecture with a ResNet50 encoder. As was the case with linear probing, we froze the weights of the encoder and trained only the decoder. We also compared our three channel RS-BYOL weights with an MS COCO based model trained on DeepLabv3 and SeCo weights to analyse the benefit of using multi-modal MS and SAR data learning with an SSL distillation network.

\subsection{Single Channel Models}

Fig. \ref{tab:evalchannelwise} shows the F1 score for the band-wise evaluation on the EuroSAT Full, EuroSAT 7K, and Sen12MS 5K datasets. To evaluate the ImageNet trained ResNet50 and BYOL model, we copied single band information into three channels, which is the standard solution available for greyscale or single band data. It is evident that RS-BYOL-1 and RS-BYOL-1 w/150K perform almost similarly, but RS-BYOL-1 w/150K outperformed all other versions in the EuroSAT full and Sen12MS 5K datasets. Especially in the SAR bands, the gap between RS-BYOL-1 w/150K and other versions is quite significant. 

With these results, it can be seen that models pre-trained in a supervised fashion with a large amount of data, such as ImageNet, show average performance. At the same time, a slight increase in remote sensing data leads to a competitive result. This opens the possibility of better representation learning in the remote sensing domain with larger datasets in a self-supervised fashion. However, the performance for a subset of EuroSAT, that is, EuroSAT 7K, shows different results, as RS-BYOL-1 outperforms RS-BYOL-1 w/150K with a performance gap of 1-3\% for most bands. To better understand if these results are significant, we also performed a t-test for the EuroSAT results, but found no significant differences between RS-BYOL-1 and RS-BYOL-1 w/150K. This emphasises that in terms of computational cost and performance, RS-BYOL-1 still outperforms all other versions.
  
As noted earlier, while BYOL and similar architectures show the best performance when trained with a very large batch size such as 4096, we only used a batch size of 32 due to hardware limitations. However, we also trained RS-BYOL-1 on a batch size of 64 to evaluate the potential impact of increasing batch sizes. However, with increasing batch size, the results did not improve compared to RS-BYOL-1 with a batch size of 32. This could be due to the small size of our training datasets in contrast to the original models trained on ImageNet. Despite the small dataset, all our single-channel versions outperformed the ImageNet trained ResNet50 models. These results predominantly demonstrate the usefulness of RS based pre-trained models.

RS-BYOL-1 trained only on MS data did not show good results compared to our MS-SAR models. In addition to that, the results with SAR bands show extreme under performance, which indicates the usefulness of multimodality for pre-training. A t-test evaluation shows p-value$<$0.05 for \textit{RS-BYOL-1 MS only} with all other variations, indicating the significant difference in overall performance. We speculate that the reason for this may be that spectral augmentation alone in remote sensing does not help the SSL model to learn invariant representations compared to using MS-SAR data.

Regarding ResNet50 and BYOL trained with ImageNet, it can be seen that bands with higher resolution (10m), i.e. Red, Green, Blue and NIR, have a smaller gap in performance compared to RS-BYOL-1, whereas bands with the lower resolution of 20m or SAR images show a more significant gap in performance with RS-BYOL-1. ResNet50 trained from scratch shows almost similar results with RS-BYOLs for EuroSAT-Full, but has similar traits to ImageNet trained ResNet50 and BYOL for EuroSAT 7K and Sen12MS. This may also be due to the small data size of EuroSAT 7K and Sen12MS, which is not surprising as training from scratch is strongly impacted by data size. This also confirms the usefulness of RS pre-trained models, as many applications such as crop monitoring or oceanography require more spectral information in the RS domain than is available in RGB images.
    
\begin{figure*}
    \centering
    \includegraphics[scale=0.5]{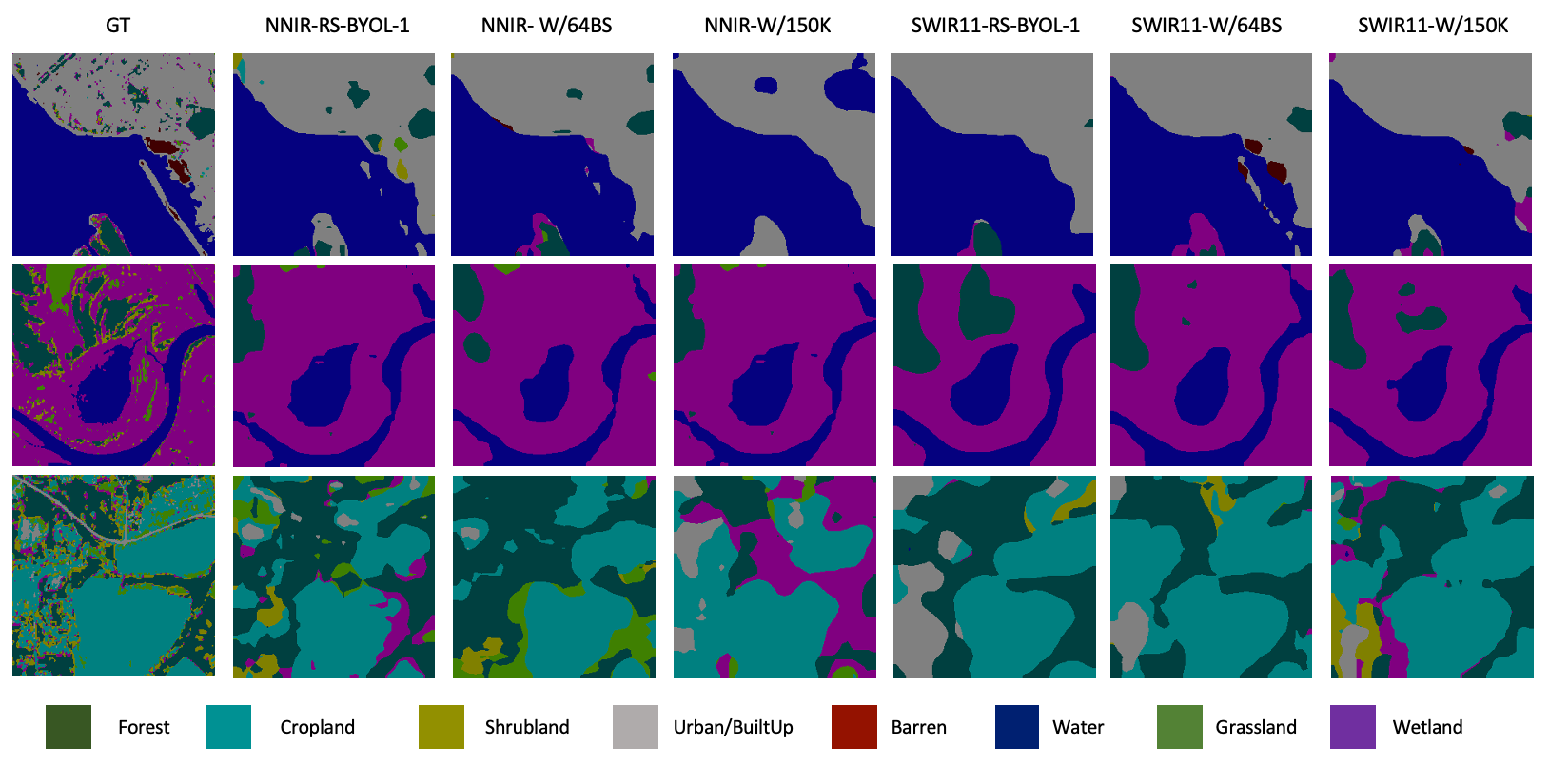}
    \caption{DFC Segmentation linear probing prediction for single channel model with NNIR and SWIR11 band. Results are skewed due to higher water pixel in the data.}
    \label{fig:seg_out}
    
\end{figure*}

Table \ref{tab:channelwisesegresults} illustrates that, for pixel-level information, increased batch size does aid performance. This can be seen in the mean intersection over union (mIoU) and Actual Accuracy (AA) values of each multi-spectral band. However, the overall difference in results in comparison to other model versions is not very large. These results also reveal that the model performs well on both SAR and MS data, as the SAR results are also competitive with MS data. It also needs to be taken into account that these results are highly skewed due to water class detection. The accuracy of the water pixels is around 90\% and above for all bands in the DFC 2020 dataset. This can be seen in Fig. \ref{fig:seg_out}, since water classes are well predicted by the three variations of RS-BYOL-1 for the NNIR and SWIR bands. This can be assumed to be due to the bands' reflection property, since NNIR and SWIR are both able to detect water pixels more profoundly. Meanwhile, the class-wise performance analysis on each band in Fig. \ref{tab:channelwisesegresults} shows that almost all bands show good results for water pixels compared to other classes. In addition, notably, almost all bands show similar performances for each class, including SAR bands.

\begin{figure*}
\centering
\begin{tabular}{cccc}
\subfloat[EuroSAT 7K on RE5 band]{\includegraphics[width=0.22\textwidth]{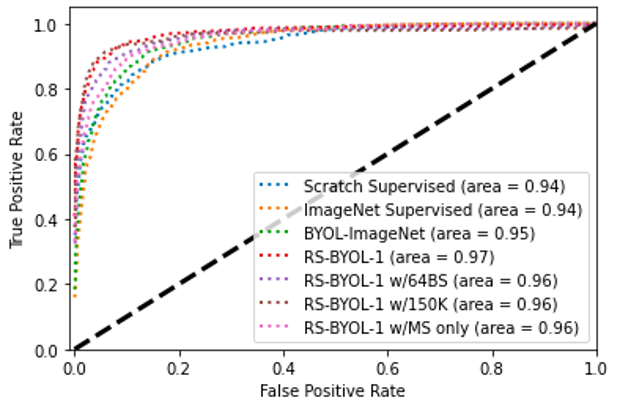}} & 
\subfloat[EuroSAT Full on RE5 band]{\includegraphics[width=0.22\textwidth]{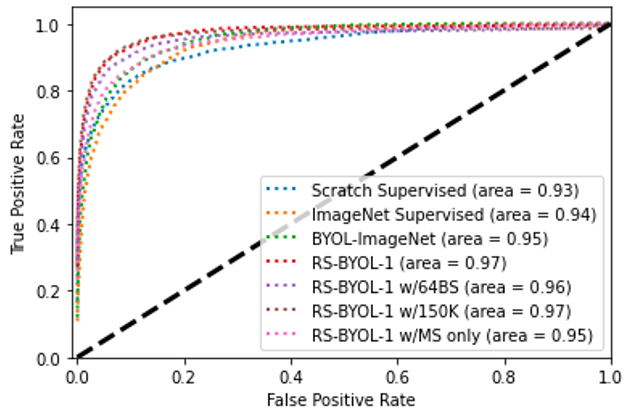}}&
\subfloat[Sen12MS on RE5 band]{\includegraphics[width=0.22\textwidth]{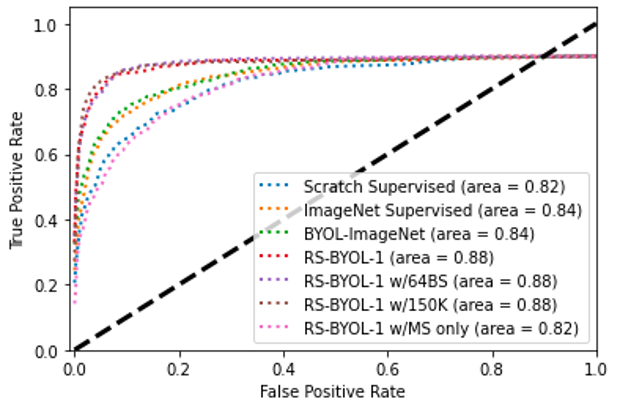}} &
\subfloat[Sen12MS on SAR (VV) band]{\includegraphics[width=0.22\textwidth]{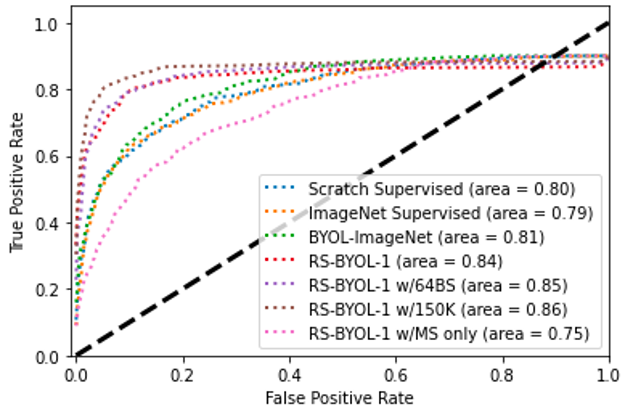}}  \\

\subfloat[EuroSAT 7K on RE5-6-7 band]{\includegraphics[width=0.22\textwidth]{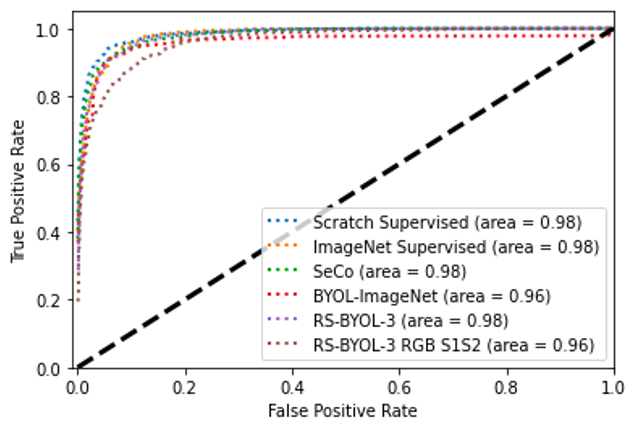}} & 
\subfloat[EuroSAT Full on RE5-6-7 band]{\includegraphics[width=0.22\textwidth]{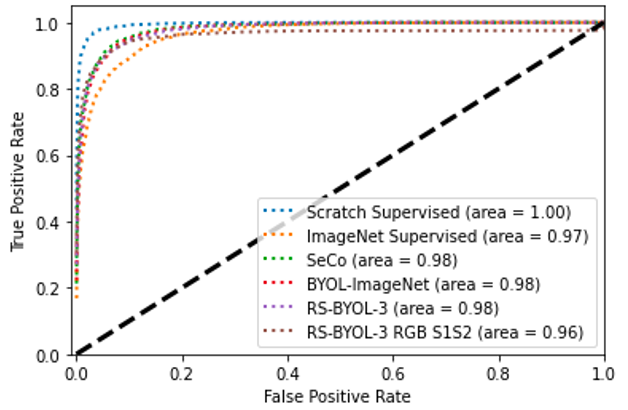}}&
\subfloat[Sen12MS on RE5-6-7 band]{\includegraphics[width=0.22\textwidth]{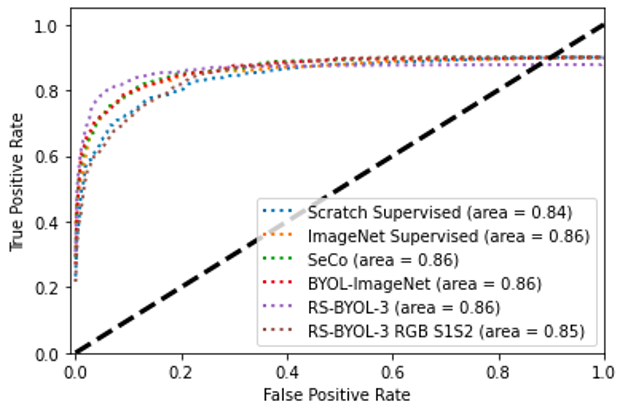}} &
\subfloat[Sen12MS on SAR (VV-VH-VV) band]{\includegraphics[width=0.22\textwidth]{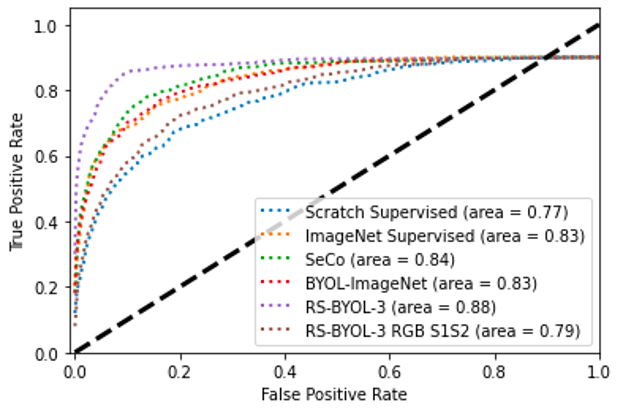}} 
 \\
\end{tabular}

\caption{ROC-AUC curve for Non-RGB bands RE5, RE5-6-7, SAR (VV) and SAR (VV-VH-VV). The comparison among models for a single channel is analogous to the results in Fig. \ref{tab:evalchannelwise}. For three channel model, as the performance gap in Table \ref{tab:evalthreechannel} for RE5-6-7 is not too large, the AUC for RS-BYOL-3 is also almost equivalent to SeCo and ImageNet trained models. However, as for SAR (VV-VH-VV), the gap is quite significant, and this can be seen in the ROC curves as well. }
\label{fig:roc_plot}
\end{figure*}

\subsection{Three Channel Models}

Table \ref{tab:evalthreechannel} presents the F1 score of our three-channel model evaluation on the EuroSAT Full, EuroSAT7K, Sen12MS and RESISC45 datasets. We compared the RS-BYOL-3 weights with the ResNet50 model trained from scratch, the ImageNet trained ResNet50, and the BYOL and SeCo weights. For all three datasets, we evaluated RS-BYOL-3's on red, green, and blue (RGB) bands, and for non-RGB bands, we utilised Red Edge bands, i.e., RE5-RE6-RE7 and SAR (VV-VH-VV) bands in the Sen12MS dataset.

\begin{table}[!htb]
\huge
\centering
\caption{Linear Probing (F1 Score) for a Three-Channel Model with RGB and Non-RGB Bands}
\resizebox{\linewidth}{!} {
\begin{tabular}{c|cc|cc|ccc|c|c} 
\hline
\multirow{2}{*}{Weights} & \multicolumn{2}{c|}{EuroSAT 7K} & \multicolumn{2}{c|}{EuroSAT Full} & \multicolumn{3}{c|}{Sen12MS}                  & RESISC45      & \begin{tabular}[c]{@{}c@{}}RESISC45\\95H\end{tabular}  \\ 
\cline{2-10}
                         & RGB           & RE5,6,7         & RGB           & RE5,6,7           & RGB           & RE5,6,7       & SAR           & RGB           & RGB                                                    \\ 
\hline
Scratch Supervised       & 0.80          & 0.86            & \textbf{0.92} & \textbf{0.91}     & 0.66          & 0.69          & 0.52          & \textbf{0.91} & 0.83                                                   \\
Random Init              & 0.46          & 0.44            & 0.50          & 0.45              & 0.40          & 0.49          & 0.27          & 0.36          & 0.27                                                   \\
ImageNet Supervised      & 0.88          & 0.87            & 0.91          & 0.88              & 0.77          & 0.77          & 0.64          & 0.90          & 0.85                                                   \\
BYOL ImageNet            & 0.88              & 0.83               & 0.89          & 0.87              & 0.78          & 0.78          & 0.66          & 0.90          & \textbf{0.88}                                          \\
SeCo \cite{manas2021seasonal}                   & \textbf{0.90} & 0.85            & \textbf{0.92}          & 0.88              & 0.81          & 0.80          & 0.68          & 0.88          & 0.82                                                   \\
RS-BYOL-3 RGB-S1S2          & 0.84          & 0.80            & 0.86          & 0.81              & 0.65          & 0.62          & 0.53          & 0.68          & 0.62                                                   \\ 

RS-BYOL-3 S1S2              & 0.88          & \textbf{0.88}   & 0.90          & 0.90              & \textbf{0.82} & \textbf{0.81} & \textbf{0.83} & 0.86          & 0.79                                                   \\

\arrayrulecolor{black}\hline
\end{tabular}
}
\label{tab:evalthreechannel}
\end{table}

\begin{table}[!htbp]
\centering
\caption{Linear probing of the three-channel model with RGB and non-RGB bands in the DFC segmentation dataset.}
\resizebox{8.5cm}{!} {
\begin{tabular}{l|cc|cc|cc|cc|cc} 
\hline
\multicolumn{1}{c|}{\multirow{2}{*}{Three Bands}} & \multicolumn{2}{c|}{\begin{tabular}[c]{@{}c@{}}CoCo \\Pre-Trained \\(Supervised)\end{tabular}} & \multicolumn{2}{c|}{RS-BYOL-3} & \multicolumn{2}{c|}{\begin{tabular}[c]{@{}c@{}}RS-BYOL-3 \\RGB Fix\end{tabular}} & \multicolumn{2}{c|}{\begin{tabular}[c]{@{}c@{}}SeCo \\RS-Trained\end{tabular}} & \multicolumn{2}{c}{\begin{tabular}[c]{@{}c@{}}BYOL \\ImageNet\end{tabular}}  \\ 
\cline{2-11}
\multicolumn{1}{c|}{}                                & mIoU & AA                                                                                      & mIoU & AA                      & mIoU & AA                                                                        & mIoU & AA                                                                      & mIoU & AA                                                                    \\ 
\hline
RGB                                               & 48.6 & 63.6                                                                                    & \textbf{51.6} & \textbf{66.9}                    & 49.7 & 64.1                                                                      & 50.2 & 64.8                                                                    & 45.0 & 59.0                                                                  \\
RE5/6/7                                           & 46.2 & 61.5                                                                                    & \textbf{51.3} & \textbf{65.8}                    & 46.2 & 61.8                                                                      & 50.5 & 64.8                                                                    & 48.0 & 62.6                                                                  \\
SAR1/2/1                                          & 44.2 & 59.9                                                                                    & \textbf{51.6} & \textbf{65.5}                    & 44.0 & 58.5                                                                      & 45.0 & 60.9                                                                    & 39.6 & 54.4                                                                  \\
\hline
\end{tabular}
}
\label{tab:threechsegresults}
\end{table}

The results in Table \ref{tab:evalthreechannel} show that ResNet50 trained from scratch performs better than pre-trained models when the size of the data set is large enough. This is not a surprise, as the size of the dataset is the limiting factor in training from scratch, leading to the rise of the concept of transfer learning. Although it can be seen that for RGB images, SeCo shows competitive results with RS-BYOLs, and also outperforms ImageNet trained ResNet50 and BYOL, except in the case of the Sen12MS datasets, where RS-BYOL-3 performed better. The good performance of ImageNet and SeCo on RGB images is not particularly surprising, as they are predominantly trained on large RGB image datasets. However, for non-RGB based images such as RE5/6/7 and SAR, RS-BYOL-3 performs better than ImageNet and SeCo trained models. This emphasises the need for RS-specific models, where the data are typically not RGB images, but MS or SAR images. In fact, it is notable that while ImageNet based models outperformed the RS based model for RGB images, the difference in results was not statistically significant and could potentially be reduced with further research on self-supervised learning.

In the case of RESISC45, the performance gap between ImageNet-supervised and RS-BYOL-3 weights is due to very high-resolution RGB data, while our RS-BYOLs are trained with low-resolution data. This particularly applies to SeCo, as their trained weights are also based on low-resolution Sentinel-2 data. However, SeCo does perform better than RS-BYOL. This evaluation suggests that SSL type learning requires more research to generalise the models across the RS domain. One solution to such a problem is to utilise both low- and very-high-resolution data for pre-training, which is also proposed by Stojnic et al. \cite{stojnic2021self}.

We argue that the under performance of \textit{RS-BYOL-3 RGB-S1S2} demonstrates that randomisation of spectral and SAR information provides better representation learning than using prominent MS data alone with the SSL distillation network.

\begin{figure}[!htbp]
    \centering
    \includegraphics[width=0.48\textwidth]{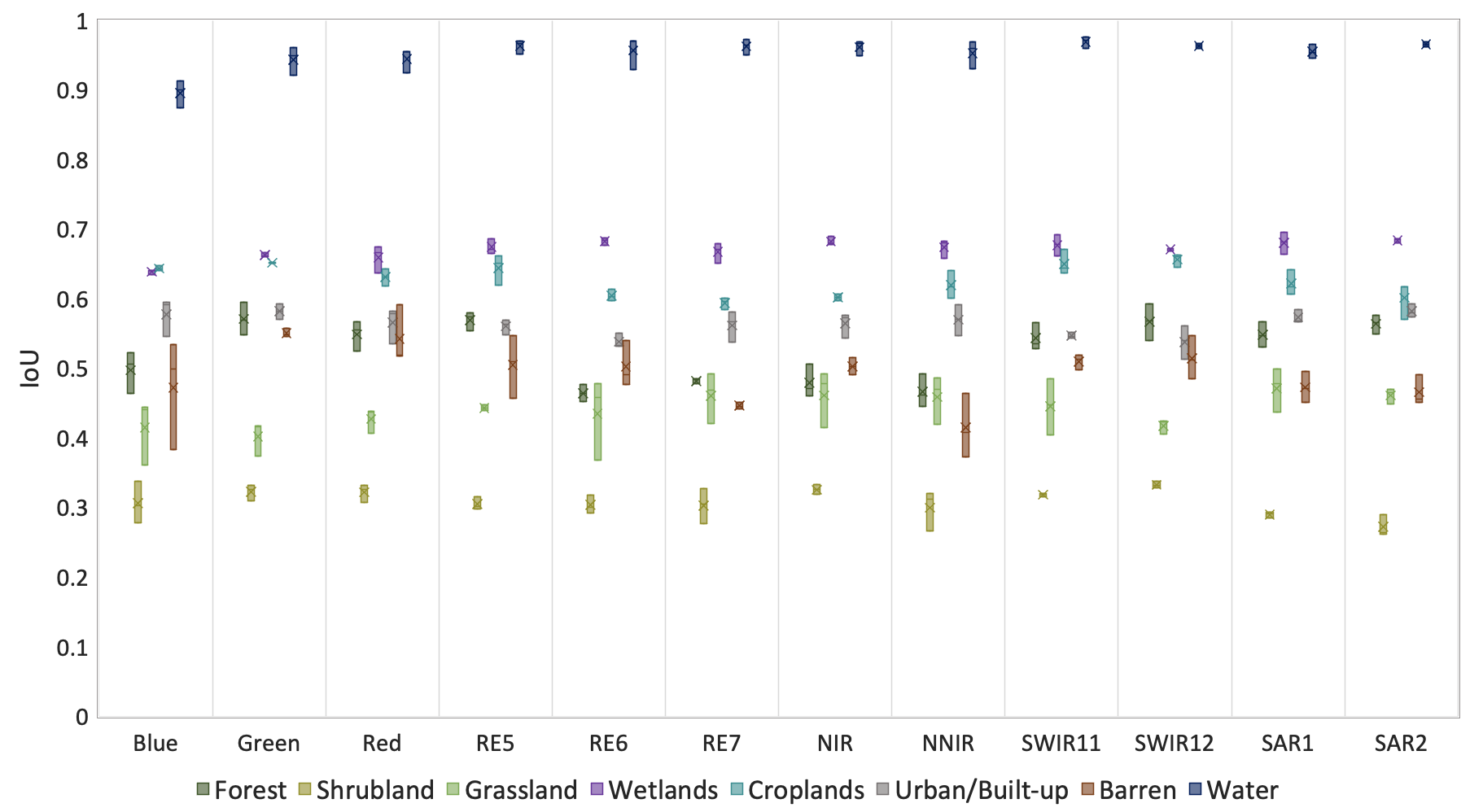}
    \caption{Class wise, performance boxplot on each band across single-channel RS-BYOLs.}
    \label{fig:seg_boundbox}
    
\end{figure}

In addition to our evaluation of learning at the pixel level using three channel models, Table \ref{tab:threechsegresults} shows the results for \textit{RS-BYOL-3} and \textit{RS-BYOL-3 RGB-S1S2}. Similar to our classification evaluation, we also compared our results with an MS COCO pre-trained model. Unlike in the classification task, the segmentation results for \textit{RS-BYOL-3} and \textit{RS-BYOL-3 RGB-S1S2} outperformed the MS COCO-based models. However, SeCo outperformed the weights of \textit{RS-BYOL-3 RGB-S1S2} and MS COCO, yet \textit{RS-BYOL-3} showed better results than SeCo. This further validates our hypothesis on the usefulness of multi-modal training in comparison to single-modal learning. The results for the segmentation remained consistent for RGB and non-RGB satellite images. However, this needs to be further evaluated on datasets other than the Sen12MS subset, as RS-BYOLs are pre-trained on the Sen12MS dataset, which might make these results skewed.

The classification and segmentation results for the single-channel models in Fig. \ref{tab:evalchannelwise} and Table \ref{tab:channelwisesegresults} report competitive performance compared to any three channel feature learning weights in Tables \ref{tab:evalthreechannel} and \ref{tab:threechsegresults}, including models pre-trained on ImageNet and MS COCO and SeCo weights. On the basis of these results, we argue that single bands can also provide good invariant representations in remote sensing, compared to the popular notion of having a model with three or more bands. 

\begin{figure*}[!htbp]
    \centering
    \includegraphics[scale=0.55]{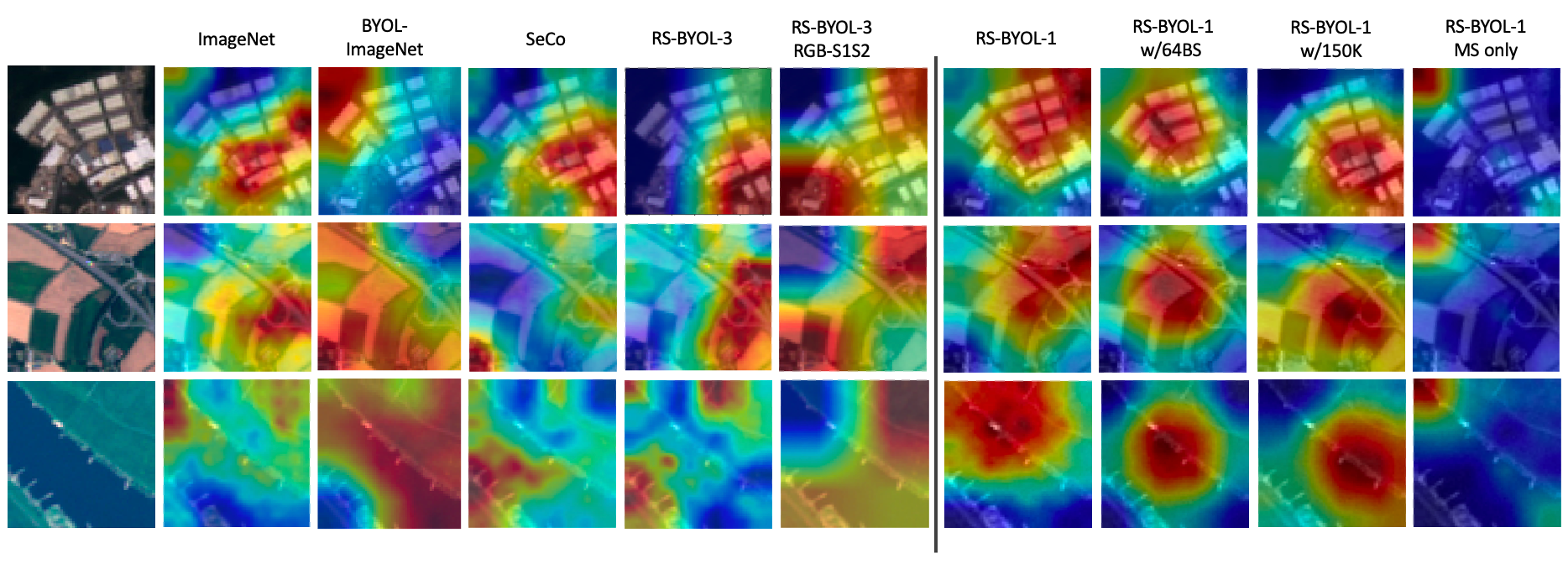}
    \caption{Gradient class activation maps \cite{srinivas2019full} for RS-BYOLs and ImageNet pre-trained ResNet50 and BYOL, and SeCo \cite{manas2021seasonal}. EuroSAT RGB images are used for three channel models, whereas Blue band image utilised for single channel models.
    Top, middle, and bottom row focuses on Industrial, Highway, and River classes respectively.}
\label{fig:rep_viz}
\end{figure*}

\begin{figure*}
\centering
\begin{tabular}{ccccc}
\subfloat[ImageNet]{\includegraphics[width=0.175\textwidth,height=3.05cm]{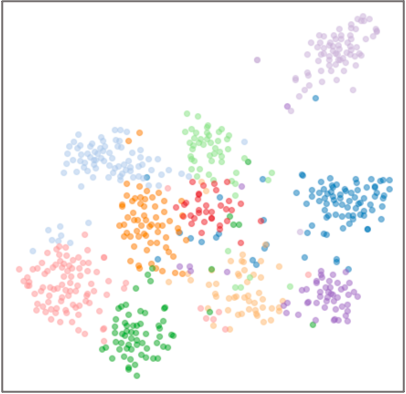}} & 
\subfloat[SeCo \cite{manas2021seasonal}]{\includegraphics[width=0.175\textwidth,height=3.05cm]{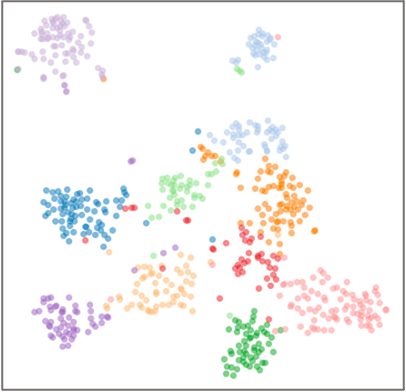}} &
\subfloat[BYOL-ImageNet-3Ch \cite{yaox12}]{\includegraphics[width=0.175\textwidth,height=3.1cm]{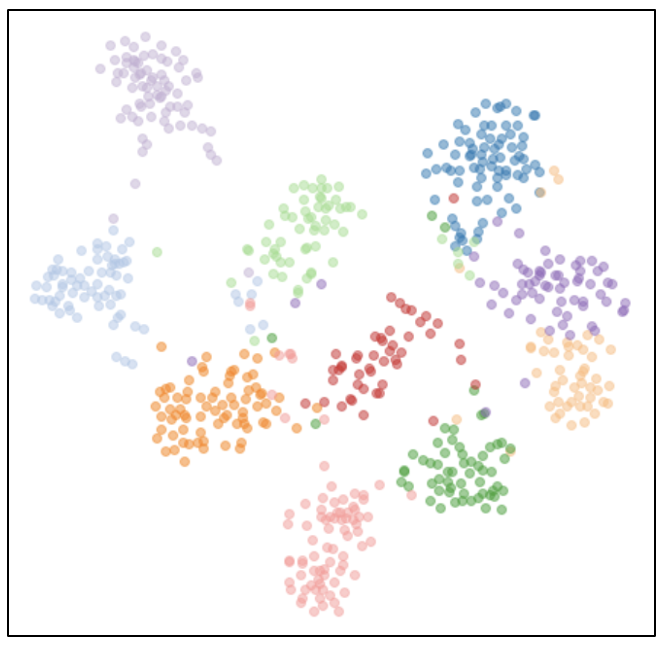}}&
\subfloat[RS-BYOL-3]{\includegraphics[width=0.175\textwidth,height=3.05cm]{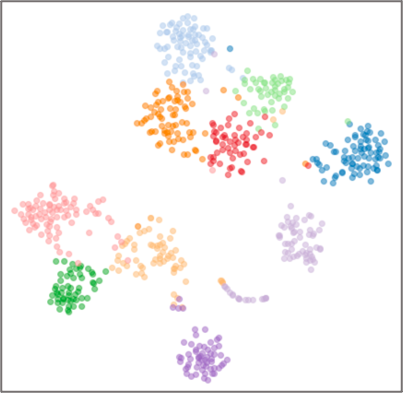}} &
\subfloat[RS-BYOL-3 RGB-S1S2]{\includegraphics[width=0.175\textwidth,height=3.05cm]{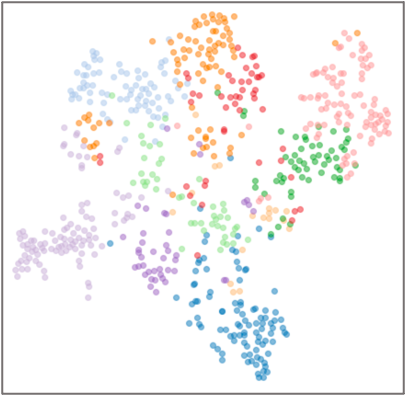}} \\
\subfloat[BYOL-ImageNet-1 \cite{yaox12}]{\includegraphics[width=0.175\textwidth,height=3.09cm]{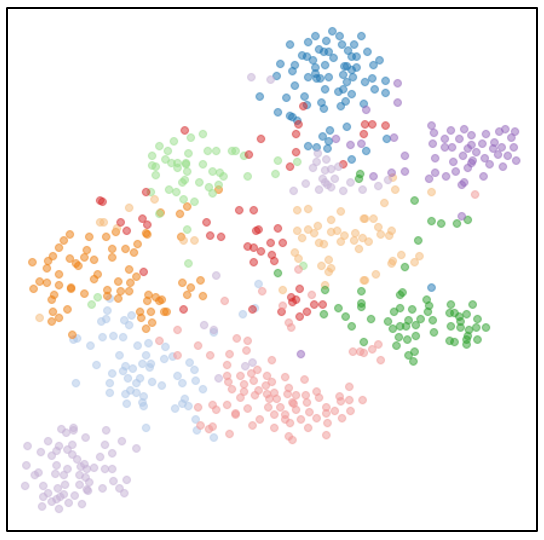}}&
\subfloat[RS-BYOL-1]{\includegraphics[width=0.175\textwidth,height=3.05cm]{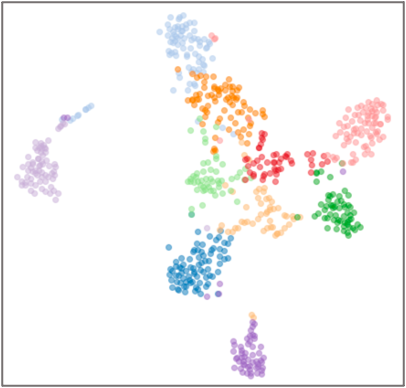}}&
\subfloat[RS-BYOL-1 w/150K]{\includegraphics[width=0.175\textwidth,height=3.05cm]{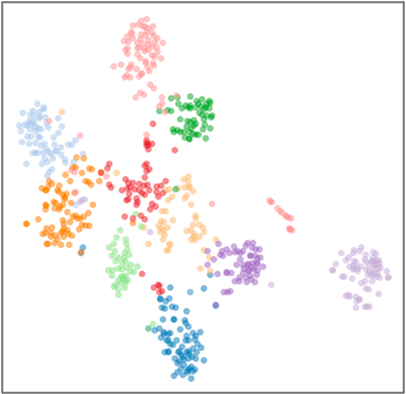}}&
\subfloat[RS-BYOL-1 w/64BS]{\includegraphics[width=0.175\textwidth,height=3.05cm]{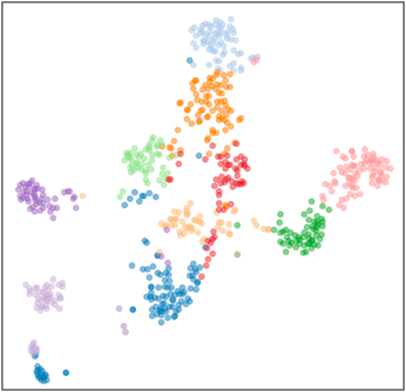}}&
\subfloat[RS-BYOL-1 MS only]{\includegraphics[width=0.175\textwidth,height=3.05cm]{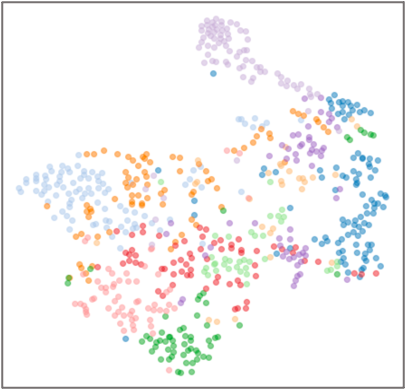}}
 \\
\end{tabular}

\caption{T-SNE plot for the EuroSAT RE5 band for single channel RS-BYOL's and RE5/6/7 bands for three channel RS-BYOLs along with ImageNet (Supervised) and BYOL, and contrastive learning based remote sensing model (SeCo) \cite{manas2021seasonal}. The Red-Edge (RE) bands are utilised to show the impact of the model on non-RGB bands.}
\label{fig:tsne_plot}
\end{figure*}

\section{Discussion}
This work proposes RS-BYOL which utilises the SSL distillation architecture (BYOL) in the RS domain with the use of multi-modality, that is, MS and SAR data. We found that multi-modality enhances the performance of the SSL based learning due to the stronger variance in data, which makes representations more invariant. The results also show that when random MS and SAR bands are utilised for single channel feature learning, it can provide competitive results compared to the common notion of three channel feature learning. 

The efficiency of RS-BYOLs are further validated by the t-SNE plots provided in Fig. \ref{fig:tsne_plot} for 10 classes of the EuroSAT 7K dataset. We compared the feature embedding from ImageNet trained ResNet50 and BYOL, SeCo, \textit{RS-BYOL-3} and \textit{RS-BYOL-3 RGB-S1S2} on RE5-6-7 band images. For single channel RS-BYOLs feature embedding, we utilised RE5 band images. It can be seen that RS-BYOLs trained with the multi-modal approach form tight clusters in the embedding space compared to ImageNet and SeCo variants. However, SeCo shows better clustering than ImageNet, which further shows the benefit of using RS based data with SSL types of learning. In addition to this, it can be seen that RS-BYOLs trained with MS dominant data did not form clean clusters on embedding. This under performance suggests that in order to learn invariant representation in low-resolution data, it is advantageous to leverage multiple modalities. 

We also looked at the full gradient class activation maps (CAM) to understand where RS-BYOLs, SeCo, and ImageNet trained ResNet50 and BYOL weights focus for class prediction in EuroSAT RGB and Blue band images. In Fig. \ref{fig:rep_viz}, it can be seen that the single channel feature embedding strongly focuses on class specific regions along with the semantic separation between two classes such as river and land area. However, CAM for single-modality \textit{RS-BYOL-1 MS only} shows out of region focus, which aligns with under performance for classification and segmentation tasks. Similar analysis can be seen in three channel MS dominant models, i.e., \textit{RS-BYOL-3 RGB-S1S2}, which shows that while making decision, its region of focus is spread across the image.

The CAM for ImageNet, SeCo, and RS-BYOL-3 weights on river class are distributed across the image and also generally fail to find the stronger region of focus except for BYOL-ImageNet. For industrial classes, all three weights have stronger focus on a smaller region of the class, despite the fact that 90\% of the image covers the class. However, in contrast, it should be noted that the same model is very precise for the highway class for ImageNet and RS-BYOL-3. While these are illustrative examples, we believe that such analysis can provide insight into the learning process. 

Both single channel and three channel RS-BYOLs performed well in downstream classification tasks, but need to improve at segmentation tasks, at least in the case of three channel RS-BYOLs. However, it should be noted that we did not evaluate these with a fine-tuning process, which might provide performance enhancement. This limitation can be further explored in the future to provide generalised network weights. 

Finally, we note that RS-based SeCo showed competitive results with RS-BYOLs, which shows the potential of SSL approaches in the RS domain to overcome the scarcity of labelled data. However, we argue that, in terms of computational efficiency, the BYOL based RS-BYOLs are more efficient than contrastive learning based. This is due to the fact that training with BYOL style networks require only positive pairs, and use of stop gradient, which in return requires less batch size and memory requirements without compromising the performance.

\section{Conclusion}
In this work, we applied the distillation network concept to build and analyse single channel and three channel features learning for MS and SAR data. MS and SAR data provide implicit augmentation with spectral and structural information differences. This work showed that using MS and SAR as different views for network training provides better representation learning than utilising MS data alone. 

We also showed that RS domain-based learning provides better results than non-RS based learning by comparing our results with results based on natural RGB datasets, such as ImageNet and MS COCO pre-trained models. Even though we trained our RS-BYOL with only 90K instances, in comparison, for example, with ImageNet (14 million), MS COCO (328K) dataset or SeCo (1 million), we achieved competitive results even with single channel feature learning alone. Another noticeable aspect of RS-BYOLs is that we trained the network with only 32 batch size, unlike other networks trained on 256 or more. This may open the path for further research work in satellite data, aiming at achieving better generalisation across the domain.

In continuation of our work, we will further explore other computationally efficient SSL networks and other RS-based pre-trained networks. Additionally, we would like to evaluate our models on further RS based datasets in order to better understand the learnt representations.


\section{Acknowledgement}
This research was conducted with the financial support of Science Foundation Ireland under Grant Agreement No. 13/RC/2106\_P2 at the ADAPT SFI Research Centre at Technological University Dublin ADAPT, the SFI Research Centre for AI-Driven Digital Content Technology, is funded by Science Foundation Ireland through the SFI Research Centres Programme.


\bibliographystyle{IEEEtran}
\bibliography{reference}

\begin{thebibliography}{10}
\providecommand{\url}[1]{#1}
\csname url@samestyle\endcsname
\providecommand{\newblock}{\relax}
\providecommand{\bibinfo}[2]{#2}
\providecommand{\BIBentrySTDinterwordspacing}{\spaceskip=0pt\relax}
\providecommand{\BIBentryALTinterwordstretchfactor}{4}
\providecommand{\BIBentryALTinterwordspacing}{\spaceskip=\fontdimen2\font plus
\BIBentryALTinterwordstretchfactor\fontdimen3\font minus
  \fontdimen4\font\relax}
\providecommand{\BIBforeignlanguage}[2]{{%
\expandafter\ifx\csname l@#1\endcsname\relax
\typeout{** WARNING: IEEEtran.bst: No hyphenation pattern has been}%
\typeout{** loaded for the language `#1'. Using the pattern for}%
\typeout{** the default language instead.}%
\else
\language=\csname l@#1\endcsname
\fi
#2}}
\providecommand{\BIBdecl}{\relax}
\BIBdecl

\bibitem{sharma2020machine}
A.~Sharma, A.~Jain, P.~Gupta, and V.~Chowdary, ``Machine learning applications
  for precision agriculture: A comprehensive review,'' \emph{IEEE Access},
  vol.~9, pp. 4843--4873, 2020.

\bibitem{ouhami2021computer}
M.~Ouhami, A.~Hafiane, Y.~Es-Saady, M.~El~Hajji, and R.~Canals, ``Computer
  vision, iot and data fusion for crop disease detection using machine
  learning: A survey and ongoing research,'' \emph{Remote Sensing}, vol.~13,
  no.~13, p. 2486, 2021.

\bibitem{zhu2020map}
Q.~Zhu, C.~Liao, H.~Hu, X.~Mei, and H.~Li, ``Map-net: Multiple attending path
  neural network for building footprint extraction from remote sensed
  imagery,'' \emph{IEEE Transactions on Geoscience and Remote Sensing},
  vol.~59, no.~7, pp. 6169--6181, 2020.

\bibitem{robinson2022fast}
C.~Robinson, A.~Ortiz, H.~Park, N.~Lozano, J.~K. Kaw, T.~Sederholm, R.~Dodhia,
  and J.~M.~L. Ferres, ``Fast building segmentation from satellite imagery and
  few local labels,'' in \emph{Proceedings of the IEEE/CVF Conference on
  Computer Vision and Pattern Recognition}, 2022, pp. 1463--1471.

\bibitem{kang2022disoptnet}
J.~Kang, Z.~Wang, R.~Zhu, J.~Xia, X.~Sun, R.~Fernandez-Beltran, and A.~Plaza,
  ``Disoptnet: Distilling semantic knowledge from optical images for
  weather-independent building segmentation,'' \emph{IEEE Transactions on
  Geoscience and Remote Sensing}, vol.~60, pp. 1--15, 2022.

\bibitem{mei2021remote}
S.~Mei, K.~Yan, M.~Ma, X.~Chen, S.~Zhang, and Q.~Du, ``Remote sensing scene
  classification using sparse representation-based framework with deep feature
  fusion,'' \emph{IEEE Journal of Selected Topics in Applied Earth Observations
  and Remote Sensing}, vol.~14, pp. 5867--5878, 2021.

\bibitem{deng2021cnns}
P.~Deng, K.~Xu, and H.~Huang, ``When cnns meet vision transformer: A joint
  framework for remote sensing scene classification,'' \emph{IEEE Geoscience
  and Remote Sensing Letters}, vol.~19, pp. 1--5, 2021.

\bibitem{li2018deepunet}
R.~Li, W.~Liu, L.~Yang, S.~Sun, W.~Hu, F.~Zhang, and W.~Li, ``Deepunet: A deep
  fully convolutional network for pixel-level sea-land segmentation,''
  \emph{IEEE Journal of Selected Topics in Applied Earth Observations and
  Remote Sensing}, vol.~11, no.~11, pp. 3954--3962, 2018.

\bibitem{khelifi2020deep}
L.~Khelifi and M.~Mignotte, ``Deep learning for change detection in remote
  sensing images: Comprehensive review and meta-analysis,'' \emph{Ieee Access},
  vol.~8, pp. 126\,385--126\,400, 2020.

\bibitem{chen2021changedetection}
Y.~Chen and L.~Bruzzone, ``Self-supervised change detection by fusing sar and
  optical multi-temporal images,'' in \emph{2021 IEEE International Geoscience
  and Remote Sensing Symposium IGARSS}.\hskip 1em plus 0.5em minus 0.4em\relax
  IEEE, 2021, pp. 3101--3104.

\bibitem{cheng2017remote}
G.~Cheng, J.~Han, and X.~Lu, ``Remote sensing image scene classification:
  Benchmark and state of the art,'' \emph{Proceedings of the IEEE}, vol. 105,
  no.~10, pp. 1865--1883, 2017.

\bibitem{gong2017feature}
M.~Gong, H.~Yang, and P.~Zhang, ``Feature learning and change feature
  classification based on deep learning for ternary change detection in sar
  images,'' \emph{ISPRS Journal of Photogrammetry and Remote Sensing}, vol.
  129, pp. 212--225, 2017.

\bibitem{pires2019convolutional}
R.~Pires~de Lima and K.~Marfurt, ``Convolutional neural network for
  remote-sensing scene classification: Transfer learning analysis,''
  \emph{Remote Sensing}, vol.~12, no.~1, p.~86, 2019.

\bibitem{ma2019deep}
L.~Ma, Y.~Liu, X.~Zhang, Y.~Ye, G.~Yin, and B.~A. Johnson, ``Deep learning in
  remote sensing applications: A meta-analysis and review,'' \emph{ISPRS
  journal of photogrammetry and remote sensing}, vol. 152, pp. 166--177, 2019.

\bibitem{wei2020improved}
Y.~Wei, X.~Luo, L.~Hu, Y.~Peng, and J.~Feng, ``An improved unsupervised
  representation learning generative adversarial network for remote sensing
  image scene classification,'' \emph{Remote Sensing Letters}, vol.~11, no.~6,
  pp. 598--607, 2020.

\bibitem{yuan2021review}
X.~Yuan, J.~Shi, and L.~Gu, ``A review of deep learning methods for semantic
  segmentation of remote sensing imagery,'' \emph{Expert Systems with
  Applications}, vol. 169, p. 114417, 2021.

\bibitem{garnot2021panoptic}
V.~S.~F. Garnot and L.~Landrieu, ``Panoptic segmentation of satellite image
  time series with convolutional temporal attention networks,'' in
  \emph{Proceedings of the IEEE/CVF International Conference on Computer
  Vision}, 2021, pp. 4872--4881.

\bibitem{persello2022deep}
C.~Persello, J.~D. Wegner, R.~Hansch, D.~Tuia, P.~Ghamisi, M.~Koeva, and
  G.~Camps-Valls, ``Deep learning and earth observation to support the
  sustainable development goals: Current approaches, open challenges, and
  future opportunities,'' \emph{IEEE Geoscience and Remote Sensing Magazine},
  2022.

\bibitem{zhu2017deep}
X.~X. Zhu, D.~Tuia, L.~Mou, G.-S. Xia, L.~Zhang, F.~Xu, and F.~Fraundorfer,
  ``Deep learning in remote sensing: A comprehensive review and list of
  resources,'' \emph{IEEE Geoscience and Remote Sensing Magazine}, vol.~5,
  no.~4, pp. 8--36, 2017.

\bibitem{schmitt2017fusion}
M.~Schmitt, F.~Tupin, and X.~X. Zhu, ``Fusion of sar and optical remote sensing
  data—challenges and recent trends,'' in \emph{2017 IEEE International
  Geoscience and Remote Sensing Symposium (IGARSS)}.\hskip 1em plus 0.5em minus
  0.4em\relax IEEE, 2017, pp. 5458--5461.

\bibitem{kulkarni2020pixel}
S.~C. Kulkarni and P.~P. Rege, ``Pixel level fusion techniques for sar and
  optical images: A review,'' \emph{Information Fusion}, vol.~59, pp. 13--29,
  2020.

\bibitem{jiang2021deep}
M.~Jiang, J.~Li, and H.~Shen, ``A deep learning-based heterogeneous
  spatio-temporal-spectral fusion: Sar and optical images,'' in \emph{2021 IEEE
  International Geoscience and Remote Sensing Symposium IGARSS}.\hskip 1em plus
  0.5em minus 0.4em\relax IEEE, 2021, pp. 1252--1255.

\bibitem{rudner2019multi3net}
T.~G. Rudner, M.~Ru{\ss}wurm, J.~Fil, R.~Pelich, B.~Bischke,
  V.~Kopa{\v{c}}kov{\'a}, and P.~Bili{\'n}ski, ``Multi3net: segmenting flooded
  buildings via fusion of multiresolution, multisensor, and multitemporal
  satellite imagery,'' in \emph{Proceedings of the AAAI Conference on
  Artificial Intelligence}, vol.~33, no.~01, 2019, pp. 702--709.

\bibitem{kang2022cfnet}
W.~Kang, Y.~Xiang, F.~Wang, and H.~You, ``Cfnet: A cross fusion network for
  joint land cover classification using optical and sar images,'' \emph{IEEE
  Journal of Selected Topics in Applied Earth Observations and Remote Sensing},
  vol.~15, pp. 1562--1574, 2022.

\bibitem{jain2021multi}
P.~Jain, B.~Schoen-Phelan, and R.~Ross, ``Multi-modal self-supervised
  representation learning for earth observation,'' in \emph{2021 IEEE
  International Geoscience and Remote Sensing Symposium IGARSS}.\hskip 1em plus
  0.5em minus 0.4em\relax IEEE, 2021, pp. 3241--3244.

\bibitem{stojnic2021self}
V.~Stojnic and V.~Risojevic, ``Self-supervised learning of remote sensing scene
  representations using contrastive multiview coding,'' in \emph{Proceedings of
  the IEEE/CVF Conference on Computer Vision and Pattern Recognition}, 2021,
  pp. 1182--1191.

\bibitem{li2021image}
Y.~Li, J.~Ma, and Y.~Zhang, ``Image retrieval from remote sensing big data: A
  survey,'' \emph{Information Fusion}, vol.~67, pp. 94--115, 2021.

\bibitem{shin2016deep}
H.-C. Shin, H.~R. Roth, M.~Gao, L.~Lu, Z.~Xu, I.~Nogues, J.~Yao, D.~Mollura,
  and R.~M. Summers, ``Deep convolutional neural networks for computer-aided
  detection: Cnn architectures, dataset characteristics and transfer
  learning,'' \emph{IEEE transactions on medical imaging}, vol.~35, no.~5, pp.
  1285--1298, 2016.

\bibitem{kim2017end}
J.~Kim and C.~Park, ``End-to-end ego lane estimation based on sequential
  transfer learning for self-driving cars,'' in \emph{Proceedings of the IEEE
  conference on computer vision and pattern recognition workshops}, 2017, pp.
  30--38.

\bibitem{sumbul2021bigearthnet}
G.~Sumbul, A.~De~Wall, T.~Kreuziger, F.~Marcelino, H.~Costa, P.~Benevides,
  M.~Caetano, B.~Demir, and V.~Markl, ``Bigearthnet-mm: A large-scale,
  multimodal, multilabel benchmark archive for remote sensing image
  classification and retrieval [software and data sets],'' \emph{IEEE
  Geoscience and Remote Sensing Magazine}, vol.~9, no.~3, pp. 174--180, 2021.

\bibitem{yuan2020self}
Y.~Yuan and L.~Lin, ``Self-supervised pre-training of transformers for
  satellite image time series classification,'' \emph{IEEE Journal of Selected
  Topics in Applied Earth Observations and Remote Sensing}, 2020.

\bibitem{vincenzi2021color}
S.~Vincenzi, A.~Porrello, P.~Buzzega, M.~Cipriano, P.~Fronte, R.~Cuccu,
  C.~Ippoliti, A.~Conte, and S.~Calderara, ``The color out of space: learning
  self-supervised representations for earth observation imagery,'' in
  \emph{2020 25th International Conference on Pattern Recognition
  (ICPR)}.\hskip 1em plus 0.5em minus 0.4em\relax IEEE, 2021, pp. 3034--3041.

\bibitem{kang2021deep}
J.~Kang, R.~Fernandez-Beltran, P.~Duan, S.~Liu, and A.~J. Plaza, ``Deep
  unsupervised embedding for remotely sensed images based on spatially
  augmented momentum contrast,'' \emph{IEEE Transactions on Geoscience and
  Remote Sensing}, vol.~59, no.~3, pp. 2598--2610, 2021.

\bibitem{grill2020bootstrap}
J.-B. Grill, F.~Strub, F.~Altch{\'e}, C.~Tallec, P.~Richemond, E.~Buchatskaya,
  C.~Doersch, B.~Avila~Pires, Z.~Guo, M.~Gheshlaghi~Azar \emph{et~al.},
  ``Bootstrap your own latent-a new approach to self-supervised learning,''
  \emph{Advances in Neural Information Processing Systems}, vol.~33, pp.
  21\,271--21\,284, 2020.

\bibitem{chen2020simple}
T.~Chen, S.~Kornblith, M.~Norouzi, and G.~Hinton, ``A simple framework for
  contrastive learning of visual representations,'' in \emph{International
  conference on machine learning}.\hskip 1em plus 0.5em minus 0.4em\relax PMLR,
  2020, pp. 1597--1607.

\bibitem{gidaris2018unsupervised}
N.~Komodakis and S.~Gidaris, ``Unsupervised representation learning by
  predicting image rotations,'' in \emph{International Conference on Learning
  Representations (ICLR)}, 2018.

\bibitem{noroozi2016unsupervised}
M.~Noroozi and P.~Favaro, ``Unsupervised learning of visual representations by
  solving jigsaw puzzles,'' in \emph{European conference on computer
  vision}.\hskip 1em plus 0.5em minus 0.4em\relax Springer, 2016, pp. 69--84.

\bibitem{zhang2016colorful}
R.~Zhang, P.~Isola, and A.~A. Efros, ``Colorful image colorization,'' in
  \emph{European conference on computer vision}.\hskip 1em plus 0.5em minus
  0.4em\relax Springer, 2016, pp. 649--666.

\bibitem{misra2020self}
I.~Misra and L.~v.~d. Maaten, ``Self-supervised learning of pretext-invariant
  representations,'' in \emph{Proceedings of the IEEE/CVF Conference on
  Computer Vision and Pattern Recognition}, 2020, pp. 6707--6717.

\bibitem{he2020momentum}
K.~He, H.~Fan, Y.~Wu, S.~Xie, and R.~Girshick, ``Momentum contrast for
  unsupervised visual representation learning,'' in \emph{Proceedings of the
  IEEE/CVF conference on computer vision and pattern recognition}, 2020, pp.
  9729--9738.

\bibitem{caron2021emerging}
M.~Caron, H.~Touvron, I.~Misra, H.~J{\'e}gou, J.~Mairal, P.~Bojanowski, and
  A.~Joulin, ``Emerging properties in self-supervised vision transformers,'' in
  \emph{Proceedings of the IEEE/CVF International Conference on Computer
  Vision}, 2021, pp. 9650--9660.

\bibitem{von2021self}
J.~Von~K{\"u}gelgen, Y.~Sharma, L.~Gresele, W.~Brendel, B.~Sch{\"o}lkopf,
  M.~Besserve, and F.~Locatello, ``Self-supervised learning with data
  augmentations provably isolates content from style,'' \emph{Advances in
  neural information processing systems}, vol.~34, pp. 16\,451--16\,467, 2021.

\bibitem{chen2021selffusion}
Y.~Chen and L.~Bruzzone, ``Self-supervised sar-optical data fusion of
  sentinel-1/-2 images,'' \emph{IEEE Transactions on Geoscience and Remote
  Sensing}, 2021.

\bibitem{wang2022self}
Y.~Wang, C.~M. Albrecht, N.~A.~A. Braham, L.~Mou, and X.~X. Zhu,
  ``Self-supervised learning in remote sensing: A review,'' \emph{arXiv
  preprint arXiv:2206.13188}, 2022.

\bibitem{manas2021seasonal}
O.~Ma{\~n}as, A.~Lacoste, X.~Giro-i Nieto, D.~Vazquez, and P.~Rodriguez,
  ``Seasonal contrast: Unsupervised pre-training from uncurated remote sensing
  data,'' in \emph{Proceedings of the IEEE/CVF International Conference on
  Computer Vision}, 2021, pp. 9414--9423.

\bibitem{zhao2020augmenting}
W.~Zhao, W.~Yamada, T.~Li, M.~Digman, and T.~Runge, ``Augmenting crop detection
  for precision agriculture with deep visual transfer learning—a case study
  of bale detection,'' \emph{Remote Sensing}, vol.~13, no.~1, p.~23, 2020.

\bibitem{qiao2021crop}
M.~Qiao, X.~He, X.~Cheng, P.~Li, H.~Luo, L.~Zhang, and Z.~Tian, ``Crop yield
  prediction from multi-spectral, multi-temporal remotely sensed imagery using
  recurrent 3d convolutional neural networks,'' \emph{International Journal of
  Applied Earth Observation and Geoinformation}, vol. 102, p. 102436, 2021.

\bibitem{akiva2021h2o}
P.~Akiva, M.~Purri, K.~Dana, B.~Tellman, and T.~Anderson, ``H2o-net:
  Self-supervised flood segmentation via adversarial domain adaptation and
  label refinement,'' in \emph{Proceedings of the IEEE/CVF Winter Conference on
  Applications of Computer Vision}, 2021, pp. 111--122.

\bibitem{bischke2019multi}
B.~Bischke, P.~Helber, J.~Folz, D.~Borth, and A.~Dengel, ``Multi-task learning
  for segmentation of building footprints with deep neural networks,'' in
  \emph{2019 IEEE International Conference on Image Processing (ICIP)}.\hskip
  1em plus 0.5em minus 0.4em\relax IEEE, 2019, pp. 1480--1484.

\bibitem{clevers2013remote}
J.~G. Clevers and A.~A. Gitelson, ``Remote estimation of crop and grass
  chlorophyll and nitrogen content using red-edge bands on sentinel-2 and-3,''
  \emph{International Journal of Applied Earth Observation and Geoinformation},
  vol.~23, pp. 344--351, 2013.

\bibitem{liu2021comprehensive}
Y.~Liu, J.~Qian, and H.~Yue, ``Comprehensive evaluation of sentinel-2 red edge
  and shortwave-infrared bands to estimate soil moisture,'' \emph{IEEE Journal
  of Selected Topics in Applied Earth Observations and Remote Sensing},
  vol.~14, pp. 7448--7465, 2021.

\bibitem{ceccato2001detecting}
P.~Ceccato, S.~Flasse, S.~Tarantola, S.~Jacquemoud, and J.-M. Gr{\'e}goire,
  ``Detecting vegetation leaf water content using reflectance in the optical
  domain,'' \emph{Remote sensing of environment}, vol.~77, no.~1, pp. 22--33,
  2001.

\bibitem{wang2006cloud}
M.~Wang and W.~Shi, ``Cloud masking for ocean color data processing in the
  coastal regions,'' \emph{IEEE Transactions on Geoscience and Remote Sensing},
  vol.~44, no.~11, pp. 3196--3105, 2006.

\bibitem{wang2013remote}
M.~Wang, S.~Son, Y.~Zhang, and W.~Shi, ``Remote sensing of water optical
  property for china's inland lake taihu using the swir atmospheric correction
  with 1640 and 2130 nm bands,'' \emph{IEEE Journal of Selected Topics in
  Applied Earth Observations and Remote Sensing}, vol.~6, no.~6, pp.
  2505--2516, 2013.

\bibitem{landuyt2018flood}
L.~Landuyt, A.~Van~Wesemael, G.~J.-P. Schumann, R.~Hostache, N.~E. Verhoest,
  and F.~M. Van~Coillie, ``Flood mapping based on synthetic aperture radar: An
  assessment of established approaches,'' \emph{IEEE Transactions on Geoscience
  and Remote Sensing}, vol.~57, no.~2, pp. 722--739, 2018.

\bibitem{zhang2018polarimetric}
L.~Zhang, Z.~Chen, B.~Zou, and Y.~Gao, ``Polarimetric sar terrain
  classification using 3d convolutional neural network,'' in \emph{IGARSS
  2018-2018 IEEE International Geoscience and Remote Sensing Symposium}.\hskip
  1em plus 0.5em minus 0.4em\relax IEEE, 2018, pp. 4551--4554.

\bibitem{klein1999sensor}
L.~A. Klein, ``Sensor and data fusion concepts and applications,'' in
  \emph{Society of Photo-Optical Instrumentation Engineers (SPIE)}, 1999.

\bibitem{zhang2020novel}
R.~Zhang, X.~Tang, S.~You, K.~Duan, H.~Xiang, and H.~Luo, ``A novel
  feature-level fusion framework using optical and sar remote sensing images
  for land use/land cover (lulc) classification in cloudy mountainous area,''
  \emph{Applied Sciences}, vol.~10, no.~8, p. 2928, 2020.

\bibitem{chen2021exploring}
X.~Chen and K.~He, ``Exploring simple siamese representation learning,'' in
  \emph{Proceedings of the IEEE/CVF Conference on Computer Vision and Pattern
  Recognition}, 2021, pp. 15\,750--15\,758.

\bibitem{zbontar2021barlow}
J.~Zbontar, L.~Jing, I.~Misra, Y.~LeCun, and S.~Deny, ``Barlow twins:
  Self-supervised learning via redundancy reduction,'' in \emph{International
  Conference on Machine Learning}.\hskip 1em plus 0.5em minus 0.4em\relax PMLR,
  2021, pp. 12\,310--12\,320.

\bibitem{tao2020remote}
C.~Tao, J.~Qi, W.~Lu, H.~Wang, and H.~Li, ``Remote sensing image scene
  classification with self-supervised paradigm under limited labeled samples,''
  \emph{IEEE Geoscience and Remote Sensing Letters}, 2020.

\bibitem{ayush2021geography}
K.~Ayush, B.~Uzkent, C.~Meng, K.~Tanmay, M.~Burke, D.~Lobell, and S.~Ermon,
  ``Geography-aware self-supervised learning,'' in \emph{Proceedings of the
  IEEE/CVF International Conference on Computer Vision}, 2021, pp.
  10\,181--10\,190.

\bibitem{montanaro2021self}
A.~Montanaro, D.~Valsesia, G.~Fracastoro, and E.~Magli, ``Self-supervised
  learning for joint sar and multispectral land cover classification,''
  \emph{arXiv preprint arXiv:2108.09075}, 2021.

\bibitem{van2020survey}
J.~E. Van~Engelen and H.~H. Hoos, ``A survey on semi-supervised learning,''
  \emph{Machine Learning}, vol. 109, no.~2, pp. 373--440, 2020.

\bibitem{richemond2020byol}
P.~Richemond, J.-B. Grill, F.~Altch{\'e}, C.~Tallec, F.~Strub, A.~Brock,
  S.~Smith, S.~De, R.~Pascanu, B.~Piot \emph{et~al.}, ``Byol works even without
  batch statistics,'' in \emph{NeurIPS 2020 Workshop: Self-Supervised
  Learning-Theory and Practice}, 2020.

\bibitem{yaox12}
\BIBentryALTinterwordspacing
X.~Yao, ``Byol-pytorch,'' 2020. [Online]. Available:
  \url{https://github.com/yaox12/BYOL-PyTorch}
\BIBentrySTDinterwordspacing

\bibitem{schmitt2019sen12ms}
M.~Schmitt, L.~H. Hughes, C.~Qiu, and X.~X. Zhu, ``Sen12ms--a curated dataset
  of georeferenced multi-spectral sentinel-1/2 imagery for deep learning and
  data fusion,'' \emph{arXiv preprint arXiv:1906.07789}, 2019.

\bibitem{helber2019eurosat}
P.~Helber, B.~Bischke, A.~Dengel, and D.~Borth, ``Eurosat: A novel dataset and
  deep learning benchmark for land use and land cover classification,''
  \emph{IEEE Journal of Selected Topics in Applied Earth Observations and
  Remote Sensing}, vol.~12, no.~7, pp. 2217--2226, 2019.

\bibitem{sen12ms}
\BIBentryALTinterwordspacing
M.~Schmitt, L.~Hughes, P.~Ghamisi, N.~Yokoya, and R.~Hansch, ``2020 ieee grss
  data fusion contest,'' \emph{IEEE Dataport}, 2019. [Online]. Available:
  \url{https://dx.doi.org/10.21227/rha7-m332}
\BIBentrySTDinterwordspacing

\bibitem{srinivas2019full}
S.~Srinivas and F.~Fleuret, ``Full-gradient representation for neural network
  visualization,'' \emph{Advances in neural information processing systems},
  vol.~32, 2019.

\end{thebibliography}

\end{document}